\documentclass{article}

\usepackage[preprint]{neurips_2026}

\usepackage[utf8]{inputenc} 
\usepackage[T1]{fontenc}    
\usepackage{hyperref}       
\usepackage{url}            
\usepackage{booktabs}       
\usepackage{amsfonts}       
\usepackage{nicefrac}       
\usepackage{microtype}      
\usepackage{xcolor}         

\usepackage{amsmath,amssymb}
\usepackage{array}
\usepackage{enumitem}
\usepackage{tabularx}
\usepackage{graphicx}

\newcolumntype{L}{>{\raggedright\arraybackslash}X}
\newcolumntype{P}[1]{>{\raggedright\arraybackslash}p{#1}}

\definecolor{darkblue}{rgb}{0, 0, 0.5}
\hypersetup{colorlinks=true, citecolor=darkblue, linkcolor=darkblue, urlcolor=darkblue,
            pdftitle={PriorZero: Bridging Language Priors and World Models for Decision Making},
            pdfauthor={Junyu Xiong, Yuan Pu, Jia Tang, Yazhe Niu}}

\title{PriorZero: Bridging Language Priors and World Models for Decision Making}

\author{
 \qquad Junyu Xiong$^{1}$ \qquad Yuan Pu$^{2}$ \qquad Jia Tang$^{3}$ \qquad Yazhe Niu$^{4,\dagger}$ \vspace{.5em}\\
$^1$University of Science and Technology of China \qquad
$^2$Shanghai Artificial Intelligence Laboratory \\
$^3$Nanjing University of Aeronautics and Astronautics \qquad
$^4$The Chinese University of Hong Kong MMLab \\
$^\dagger$Corresponding author.
}

\begin{document}

\maketitle

\vspace{-18pt}
\begin{abstract}
Leveraging the rich world knowledge of Large Language Models (LLMs) to enhance Reinforcement Learning (RL) agents offers a promising path toward general intelligence.
However, a fundamental prior-dynamics mismatch hinders existing approaches: static LLM knowledge cannot directly adapt to the complex transition dynamics of long-horizon tasks.
Using LLM priors as fixed policies limits exploration diversity, as the prior is blind to environment-specific dynamics; while end-to-end fine-tuning suffers from optimization instability and credit assignment issues.
To bridge this gap, we propose \textit{PriorZero}, a unified framework that integrates LLM-derived conceptual priors into world-model-based planning through a decoupled rollout-training design.
During rollout, a novel root-prior injection mechanism incorporates LLM priors exclusively at the root node of Monte Carlo Tree Search (MCTS), focusing search on semantically promising actions while preserving the world model's deep lookahead capability.
During training, PriorZero decouples world-model learning from LLM adaptation: the world model is continuously refined on interaction data to jointly improve its dynamics, policy, and value predictions, its value estimates are then leveraged to provide fine-grained credit assignment signals for stable LLM fine-tuning via alternating optimization.
Experiments across diverse benchmarks, including text-based adventure games in Jericho and instruction-following gridworld tasks in BabyAI, demonstrate that PriorZero consistently improves both exploration efficiency and asymptotic performance, establishing a promising framework for LLM-empowered decision-making.
Our code is available at \textcolor{magenta}{https://github.com/opendilab/LightZero}.
\end{abstract}

\vspace{-5pt}
\section{Introduction}
\vspace{-3pt}
\label{sec:intro}

Large Language Models (LLMs) encode broad semantic and commonsense knowledge that can significantly benefit reinforcement learning (RL) agents in language-conditioned, embodied, and interactive decision-making tasks~\citep{huang2022language,ahn2022can,li2022pre,carta2023grounding}. Such knowledge is particularly valuable in long-horizon environments with sparse rewards, where purely trial-and-error exploration often entails prohibitive interaction costs.

However, transferring the static, declarative knowledge of LLMs to dynamic RL tasks remains challenging. The core difficulty lies in a prior--dynamics mismatch: LLMs excel at semantic plausibility and high-level reasoning, whereas RL agents must model environment-specific transitions, assign long-term credit, and plan under feedback. Existing paradigms expose this mismatch in different ways. Directly using an LLM as a policy~\citep{huang2022language,ahn2022can,huang2023inner} often leads to myopic decisions because linguistic likelihood is not equivalent to environment return. End-to-end RL fine-tuning of LLM policies~\citep{carta2023grounding} introduces environment feedback, but suffers from high-variance credit assignment and poor scalability in long-horizon sparse-reward tasks. Using language models as dynamics models~\citep{lin2024dynalang} can leverage language-conditioned prediction, but accurate action-conditioned transition modeling remains difficult: LLMs are pretrained via token-level autoregressive objectives, whereas latent dynamics models must predict state transitions at the state-action level---this granularity gap prevents the pretrained knowledge from being directly reused for dynamics learning. 
Finally, treating LLMs as frozen representations~\citep{li2022pre} improves sample efficiency but limits task-specific adaptation.

To address these bottlenecks, we propose PriorZero, a unified framework that integrates LLM semantic priors into world-model-based latent planning. Building upon recent advancements in model-based RL and latent world models~\citep{hansen2024tdmpc2,wang2024efficientzero,unizero,xuan2024rezero,niu2024lightzero,pu2025one}, PriorZero places a learnable world model between the LLM and the environment. This design enables the LLM to provide valuable semantic guidance while delegating transition modeling, value estimation, and multi-step planning to the world model. PriorZero decouples the rollout and training processes through two complementary mechanisms.

During rollout, we introduce \textit{Root-Prior Injection}: LLM semantic priors are fused exclusively at the root node of latent-space Monte Carlo Tree Search (MCTS). This provides the search with an informative initial direction while preserving the world model's capacity to conduct deeper latent exploration, effectively preventing MCTS from falling into meaningless trajectory exploration (e.g., reward loops or logically inconsistent paths).

During training, we introduces \textit{Alternating Reinforcement Fine-Tuning}, which decouples world-model learning from LLM adaptation. First, the interaction data collected under the guidance of LLM priors is used to update the world model, accelerating the convergence of its components. Then, the world model's action-value estimates are utilized to construct 
fine-grained credit assignment signals for stable LLM fine-tuning. This alternating process aligns the LLM prior more closely with environment-specific decision outcomes while avoiding the instability of end-to-end optimization.

Experiments on Jericho~\cite{hausknecht2020jericho} and BabyAI~\cite{chevalier2019babyai} demonstrate that PriorZero improves exploration efficiency and final performance across text-based adventure games and  gridworld instruction-following tasks. Our main contributions are:
\begin{itemize}[leftmargin=*]
    \item We analyze the major paradigms for transferring LLM knowledge to RL agents and identify the \textit{prior-dynamics mismatch} as the central obstacle in long-horizon, sparse-reward decision-making.
    \item We propose PriorZero, a framework that synergizes LLM priors with world-model planning through two core mechanisms: \textit{Root-Prior Injection}, which injects semantic guidance at the MCTS root node to facilitate exploration; and \textit{Alternating Reinforcement Fine-Tuning}, which leverages world-model value estimates as fine-grained credit assignment signals for stable LLM adaptation.
    \item We demonstrate strong performance on the Jericho and BabyAI benchmarks, and validate the specific contributions of root-prior injection, LLM fine-tuning, CoT-based prior extraction, model scale, and MCTS through comprehensive ablations.
\end{itemize}
\section{Categorization of LLM Knowledge Transfer Paradigms}
\label{sec:motivation}

\begin{figure}[t]
    \centering
    \includegraphics[width=0.75\textwidth]{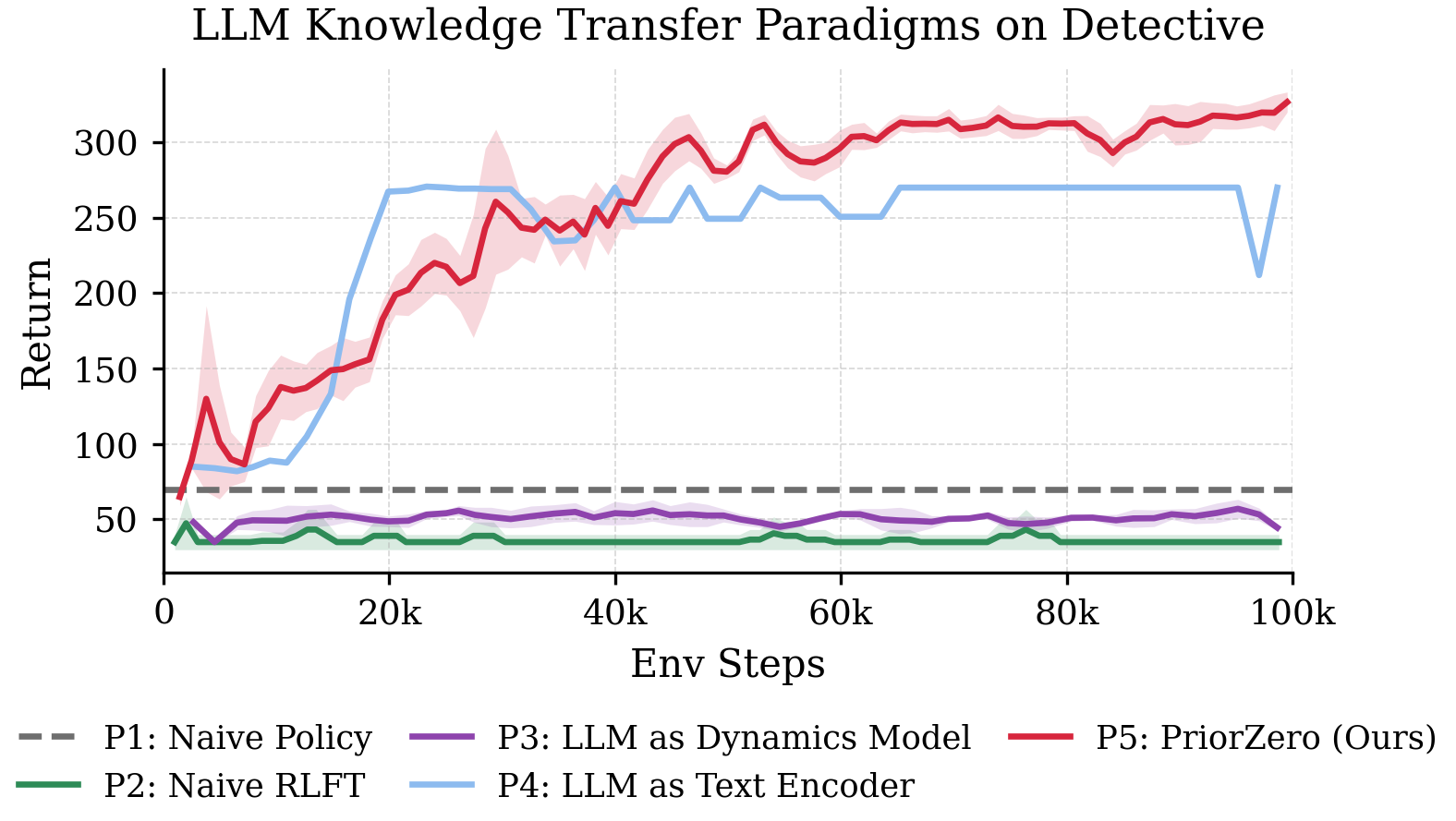}
      \caption{Training curves of five LLM knowledge transfer paradigms on the \texttt{Detective} environment.
      P1--P3 stagnate at low return due to their structural bottlenecks: myopic action selection (P1), high-variance end-to-end updates (P2), and granularity mismatch between token-level pretraining and            state-action-level dynamics (P3). P4 improves steadily but saturates below PriorZero due to its non-adaptive text representation. PriorZero achieves both faster convergence and the highest asymptotic return through its decoupling design (Table~\ref{tab:paradigms_comparison}).} 
    \label{fig:paradigms_curves}
    \vspace{-12pt}
\end{figure}

Transferring the world knowledge of LLMs to RL agents is a compelling yet challenging research direction~\citep{shi2025monte,yu2026rwml}.
We categorize existing approaches~\citep{huang2022language, ahn2022can,huang2023inner, szot2024llarp, lin2024dynalang, shi2025monte, yu2026rwml} along three dimensions---LLM plasticity, core mechanism, and primary bottleneck---and identify established paradigms (Table~\ref{tab:paradigms_comparison}).
Figure~\ref{fig:paradigms_curves} compares their performance on a representative Jericho task (Detective). Detailed analysis of each paradigm are provided in Appendix~\ref{sec:appendix_paradigms}.

\begin{table}[t]
\centering
\caption{Comparison of LLM knowledge transfer paradigms for RL, characterized in LLM plasticity, core mechanism, and primary bottleneck. Existing paradigms (P1--P4) each face a distinct structural bottleneck in long-horizon, sparse-reward tasks. PriorZero resolves these by decoupling the LLM prior from world-model dynamics, yielding a favorable balance between adaptability and stability.}
\label{tab:paradigms_comparison}
\footnotesize
\renewcommand{\arraystretch}{1.25}
\setlength{\tabcolsep}{4pt}
\begin{tabularx}{\textwidth}{%
  >{\raggedright\arraybackslash}p{2.0cm}
  c
  >{\raggedright\arraybackslash}X
  >{\raggedright\arraybackslash}X}
\toprule
\textbf{Paradigm} & \textbf{LLM Plasticity} & \textbf{Core Mechanism} & \textbf{Primary Bottleneck} \\
\midrule
Naive Policy
  & Frozen
  & Prompts LLM to directly output actions in text format.
  & \textit{Knowing-Doing Gap}: no environment-specific adaptation. \\
Naive RLFT
  & Trained
  & LLM as learnable policy, end-to-end RL Training.
  & Credit assignment failure under sparse rewards and diverse feedbacks. \\
Dynamics Model
  & Trained
  & LLM predicts state transitions.
  & \textit{Granularity Mismatch}: token-level $\neq$ state-action-level. \\
Text Encoder
  & Frozen
  & LLM encodes texts and extracts static semantic features.
  & \textit{Representation Bottleneck}: non-adaptive. \\
\midrule
\textbf{PriorZero}
  & Alternating
  & Decouples LLM from world-model; closed-loop mutual improvement.
  & Requires balancing LLM prior and world-model policy. \\
\bottomrule
\end{tabularx}
\end{table}

\textbf{P1 (Naive Policy)} directly prompts the LLM to select actions~\citep{huang2022language,ahn2022can,wang2023voyager,yao2022react}, but the absence of an environmental feedback loop yields myopic policies---the ``knowing-doing gap''.
\textbf{P2 (Naive RLFT)} fine-tunes the LLM end-to-end via RL~\citep{schmied2023l2m,zhai2024fine,carta2023grounding}, yet suffers from high-variance gradients under sparse rewards, poor exploration in large action spaces, and prohibitive computational cost.
\textbf{P3 (Dynamics Modeling)} uses the LLM to predict state transitions~\citep{hao2023reasoning,xie2025making}, but a granularity mismatch between token-level text generation and state-action-level transition prediction
 prevents the LLM's pretrained knowledge from being effectively reused for latent dynamics learning.
\textbf{P4 (Text Encoder)} extracts static LLM features for downstream RL modules~\citep{xu2025vlms,szot2024llarp,wang2024llm,yan2025efficient}, avoiding catastrophic forgetting but imposing a performance ceiling due to non-adaptive representations.
These limitations motivate PriorZero's design: rather than forcing the LLM into a role in decision-making tasks, which is often ill-suited (e.g., dynamics modeling or end-to-end policy), we let it provide what it does best---semantic priors---and delegate dynamics and planning to a world model.
\vspace{-8pt}
\section{PriorZero}
\vspace{-4pt}
\label{sec:priorzero}

\begin{figure}[t]
    \centering
    \includegraphics[width=0.9\textwidth]{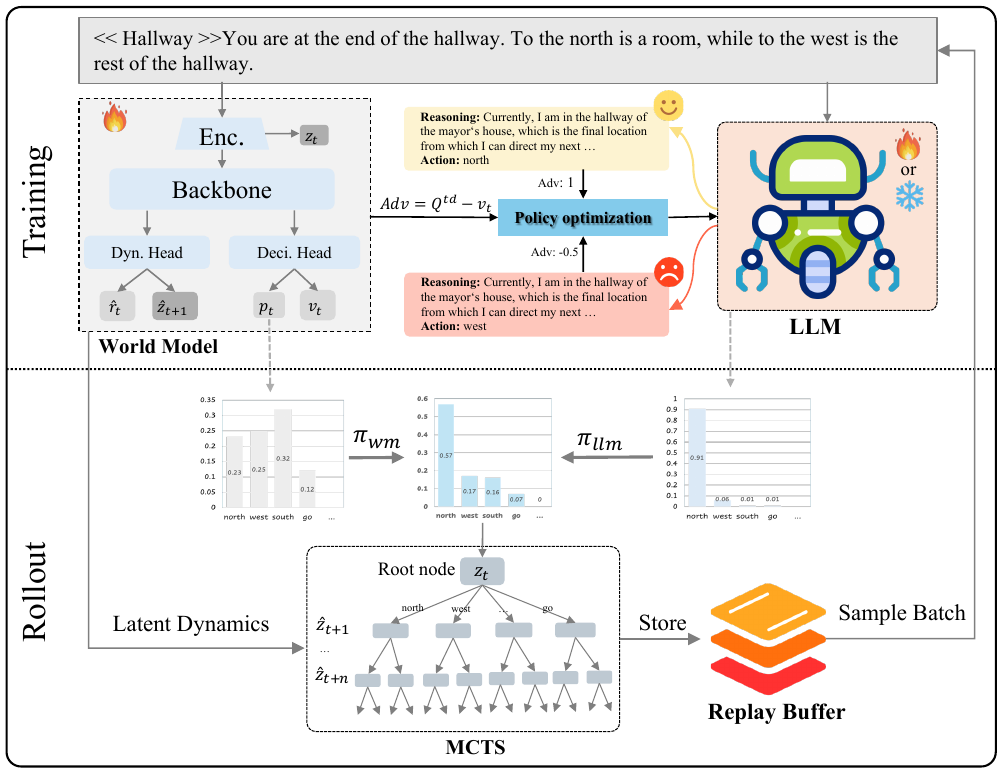}
    \caption{Overview of PriorZero. \textbf{Bottom}: during rollout, the LLM produces a semantic action prior $\pi_{\mathrm{LLM}}$ via chain-of-thought reasoning, which is fused with the world-model policy $\pi_{\mathrm{WM}}$ \emph{exclusively at the MCTS root node}; non-root nodes rely entirely on the world model. \textbf{Top}: during training, the world model is updated on interaction data to refine its dynamics, value, and policy predictions; the resulting value estimates construct 
    low-variance advantage signals for alternating LLM fine-tuning via PPO, forming a closed-loop mutual-improvement cycle.}
    \label{fig:framework}
    \vspace{-12pt}
\end{figure}

PriorZero proposes a decoupled rollout-training architecture that integrates Large Language Model (LLM) semantic priors into a world-model-based planning framework (Figure~\ref{fig:framework}).
Two principles guide its design, each directly responding to a failure mode identified in Section~\ref{sec:motivation}: \textit{(i)~the LLM prior must not interfere with the world model's faithful dynamics modeling}---preventing the granularity mismatch that disabled P3; and \textit{(ii)~the world model's planning knowledge must provide a low-variance, fine-grained supervision signal for LLM adaptation}---resolving the credit-assignment failure that plagued P2.
During rollout, LLM-derived action priors are injected exclusively at the MCTS root node (\S\ref{subsec:root_injection}).
During training, world-model learning and LLM adaptation are optimized under distinct objectives in an alternating schedule (\S\ref{subsec:decoupled_training}).
Full implementation details are in Appendix~\ref{sec:appendix_implementation}.

\subsection{Planning with a Latent World Model}
\label{subsec:latent_planning}
\vspace{-10pt}
PriorZero adopts the same world-model setup as UniZero~\citep{unizero}: a unified Transformer that processes interleaved observation and action embeddings to produce context-conditioned latent representations. Let $\mathcal{H}_t = (o_{t-H+1}, a_{t-H+1}, \ldots, o_t)$ denote the historical context of length $H$ at step $t$ ($H{=}10$ during training and $H{=}4$ at test time). 
The Transformer encodes this sequence into $z_t = h_\theta(\mathcal{H}_t)$, from which prediction heads output the policy and value:
\[
\pi_{\mathrm{WM}}(\cdot \mid z_t) = f_\theta^{\pi}(z_t), \qquad v_\theta(z_t) = f_\theta^{v}(z_t).
\]
Unlike LLMs, which model long-range dependencies at the token level, the world model captures long-range dependencies at the \emph{state-action level}, directly relevant for credit assignment and planning.
During the MCTS procedure, the dynamics model recursively unrolls imagined trajectories in the latent space, starting from $z_t$:
\[
(\hat{z}_{t+k+1}, \hat{r}_{t+k}) = g_\theta(\hat{z}_{t+k}, a_{t+k}), \quad k = 0, 1, \ldots,
\]
where $\hat{z}_t \equiv z_t$. In parallel, the same history is serialized into a textualized prompt $\mathcal{C}_t$ (the language counterpart of $\mathcal{H}_t$) together with the candidate action set $\mathcal{A}_t$. The pretrained LLM then serves as an independent commonsense oracle, producing an action prior $\pi_{\mathrm{LLM}}(\cdot \mid \mathcal{C}_t)$ that encodes its plausibility judgment over $\mathcal{A}_t$, obtained through chain-of-thought reasoning (detailed below).

\subsection{Root-Prior Injection}
\label{subsec:root_injection}

\paragraph{Why MCTS?}
A pretrained LLM offers strong priors but no environment-specific lookahead; pure model-free RL lacks sample-efficient credit assignment under sparse rewards. MCTS provides a step-wise policy-improvement operator over a learned latent dynamics model, turning $\pi_{\mathrm{WM}}$ at each step into a stronger search-improved distribution. The remaining question is \emph{how to inject the LLM prior into MCTS without contaminating downstream simulation}.

\paragraph{Root-Only Injection.}
The core design of PriorZero is to fuse the LLM prior \textit{exclusively} at the MCTS root node. The root corresponds to $z_t$, which encodes the full historical context $\mathcal{H}_t$ and aligns with the LLM's judgment based on $\mathcal{C}_t$. Non-root nodes in the search tree are recursively generated by the latent dynamics model and have no direct textual counterpart; injecting the LLM prior at such nodes would force a textual interpretation onto a state that no longer admits one, introducing systematic bias.
The fused prior at the root and the world-model-only prior at non-root nodes are:
\[
P_{\mathrm{root}}(a \mid z_t) = (1 - \alpha)\,\pi_{\mathrm{WM}}(a \mid z_t) + \alpha\,\pi_{\mathrm{LLM}}(a \mid \mathcal{C}_t),
\qquad
P_{\mathrm{non\text{-}root}}(a \mid \hat{z}) = \pi_{\mathrm{WM}}(a \mid \hat{z}).
\]
Node selection follows the standard pUCT criterion. The MCTS lookahead resolves the myopia of LLM-only policies (P1) by replacing single-step likelihood with multi-step value estimation. Upon completion of the search, the root visit counts are temperature-scaled into the execution policy $\pi_{\mathrm{MCTS}}(a \mid z_t) \propto N(z_t, a)^{1/\tau}$.

\paragraph{Structured Prior Extraction.}
To obtain high-quality priors, we use a two-stage prompt template. 
The LLM first produces a situation analysis $c_t$ under a \emph{neutrality constraint}---a system instruction that requires the analysis to describe the state objectively without committing to any action (see example in Appendix~\ref{sec:appendix_implementation}). Conditioned on $(\mathcal{C}_t, c_t)$, the LLM then outputs the action distribution:
\[
\pi_{\mathrm{LLM}}(a \mid \mathcal{C}_t) \triangleq P_{\mathrm{LLM}}(a \mid \mathcal{C}_t, c_t),
\quad c_t \sim \mathrm{LLM}(\cdot \mid \mathcal{C}_t, \texttt{[neutrality\_constraint]}).
\]
This ``analyze-before-decide'' design avoids the \emph{rationalization bias} where the LLM, having committed to an action, retroactively constructs reasoning that justifies it. Root-prior injection thus achieves a clean division of labor: the LLM \emph{biases} the search toward semantically promising actions, while the world model performs the multi-step lookahead that determines their actual long-term value.

\subsection{Decoupled Training Strategy}
\label{subsec:decoupled_training}


\paragraph{Alternating Fine-Tuning.}
World-model learning and LLM policy fine-tuning use distinct optimization objectives over a shared interaction stream, forming a closed-loop training pipeline.
The two modules follow an asymmetric alternating schedule: $N_{\mathrm{WM}}$ world-model updates are followed by $N_{\mathrm{LLM}}$ LLM updates, with $N_{\mathrm{WM}}{:}N_{\mathrm{LLM}}{=}10{:}1$. The world-model objective follows UniZero~\citep{unizero}, including policy distillation, distributional value prediction (categorical head over a discretized support), reward prediction, and latent consistency losses.
A world-model warm-up phase precedes the first LLM update to ensure that the value is sufficiently accurate to provide meaningful signal.

The core innovation is to construct a low-variance policy-gradient signal for the LLM from the world model's value estimates. We use the $n$-step bootstrapped TD advantage:
\[
A(z_t, a_t) = Q_n(z_t, a_t) - v_\theta(z_t),
\quad Q_n(z_t, a_t) = \sum_{i=0}^{n-1} \gamma^i r_{t+i} + \gamma^n v_\theta(z_{t+n}),
\]
where $v_\theta$ is the history-conditioned world-model value head, trained against MCTS-improved targets. Combining $n$ real rewards with a trained bootstrap keeps variance well below Monte-Carlo returns.
Raw advantages are normalized to $\hat{A}$ within each LLM phase, then softly fused with a binary structured-template reward $r_{\mathrm{fmt}}$ to obtain $\hat{A}_{\mathrm{final}}$ (blending coefficient in Appendix~\ref{sec:appendix_implementation}); this enforces schema compliance without overwhelming the value signal. The LLM parameters $\phi$ are updated via the clipped PPO objective with a KL regularizer to a frozen reference $\pi_{\mathrm{ref}}$, following the RLFT recipes in LightRFT~\citep{lightrft} and OpenRLHF~\citep{hu2024openrlhf}:
\[
\mathcal{L}_{\mathrm{total}}(\phi) = \underbrace{-\,\mathbb{E}_t\!\left[\min\!\big(\rho_t\,\hat{A}_{\mathrm{final}},\,\mathrm{clip}(\rho_t, 1{-}\epsilon, 1{+}\epsilon)\,\hat{A}_{\mathrm{final}}\big)\right]}_{\mathcal{L}_{\mathrm{PPO}}} \;+\; \beta\,D_{\mathrm{KL}}(\pi_\phi \,\|\, \pi_{\mathrm{ref}}),
\]
where $\rho_t = \pi_\phi(y_t\mid\cdot)/\pi_{\phi_{\mathrm{old}}}(y_t\mid\cdot)$ is the per-token importance ratio over the generated CoT-and-action sequence. To focus updates on the tokens that determine the executed action, we scale the per-token loss of the CoT prefix by $w_{\mathrm{cot}}{=}0.1$ while action tokens carry full weight; the ratio $\rho_t$ itself is unaffected.

\paragraph{Why This Decoupling Works.}
The closed loop directly addresses each failure mode identified in Section~\ref{sec:motivation}:
(P1)~the \emph{knowing-doing gap} is closed because the LLM's prior is grounded by world-model lookahead at every step;
(P2)~the \emph{credit-assignment failure} of end-to-end RLFT is avoided because gradients reach the LLM only through the world-model's variance-reduced advantage, never through long on-policy returns;
(P3)~the \emph{granularity mismatch} is sidestepped because the LLM is never asked to predict transitions---it provides priors only at the root, where textual and latent representations align;
(P4)~the \emph{representation bottleneck} is broken because the LLM is trainable, allowing its priors to co-evolve with the world model.
The overall effect is a progressive alignment between LLM-encoded world knowledge and the environment's state-action-level dynamics.

\section{Experiments and Analysis}
\label{sec:experiments}
\vspace{-3pt}
\subsection{Experimental Setup}
\vspace{-2pt}

\paragraph{Environments.}
We evaluate PriorZero on two complementary benchmarks.
The primary evaluation uses \textbf{Jericho}~\cite{hausknecht2020jericho}, a suite of text-based adventure games that stress long-horizon planning under sparse rewards.
All methods restrict their action selection to the \texttt{valid\_actions} provided by Jericho.
To test generalization beyond text games, we further evaluate on \textbf{BabyAI}~\cite{chevalier2019babyai}, a grid-world environment with natural-language commands (\S\ref{subsec:babyai}).
For Jericho, we select four games spanning distinct capability dimensions:
\begin{itemize}[leftmargin=*]
    \item \textbf{Detective}---a mystery-solving game with short optimal paths ($<$100 steps) and compact action spaces ($<$10 per step). It tests \textit{commonsense reasoning} via logical inference from textual clues.

    \item \textbf{Acorncourt}---a puzzle-driven adventure with moderate horizons. Evaluates \textit{combinatorial action-space handling}, requiring objects to be discovered and combined in the correct sequence.

    \item \textbf{Zork1}---a dungeon-exploration game with extended trajectories ($>$300 steps) and over 50 actions per step. Evaluates \textit{long-horizon planning} under sparse rewards across a vast, interconnected map.

    \item \textbf{Omniquest}---a fantasy adventure with limited environmental feedback. Evaluates \textit{sparse-reward exploration}, as meaningful rewards are separated by long sequences of unrewarded actions.
\end{itemize}

\paragraph{Implementation Details.}
Our experiments build upon the unified MCTS framework established by LightZero~\citep{niu2024lightzero}.
We instantiate PriorZero on top of UniZero~\citep{unizero} with \texttt{Qwen2.5-3B-Instruct} as the LLM. During inference, the LLM scores candidate actions conditioned on a chain-of-thought prefix; log-probabilities are normalized via softmax with temperature 1 to produce a smooth policy prior.
During training, the fusion weight is $\alpha{=}0.5$, the CoT loss weight is $w_{\mathrm{cot}}{=}0.1$, and the alternating schedule performs $N_{\mathrm{WM}}{=}2000$ world-model steps followed by $N_{\mathrm{LLM}}{=}200$ LLM PPO steps (AdamW, lr $1{\times}10^{-6}$, KL weight $0.01$). Full hyperparameters are provided in Appendix~\ref{sec:appendix_implementation}, and a per-paradigm compute breakdown is reported in Appendix~\ref{sec:appendix_implementation}.
\vspace{-5pt}

\paragraph{Evaluation Metrics.}
We compare PriorZero with baselines in terms of convergence behavior across different environments, assessing overall decision-making capability and learning efficiency. In addition, we analyze the entropy of the MCTS root-node visit distributions, which reflects policy uncertainty and provides insight into how semantic priors influence exploration.\footnote{The main comparison runs use three random seeds. Failed-variant ablation curves are reported from a single seed: preliminary runs already established their non-viability, and additional seeds would not have changed the qualitative conclusion.}

\vspace{-5pt}
\subsection{Main Results on Jericho}
\label{subsec:main_results}
\vspace{-5pt}

\begin{figure}[t]
    \centering
    \includegraphics[width=\textwidth]{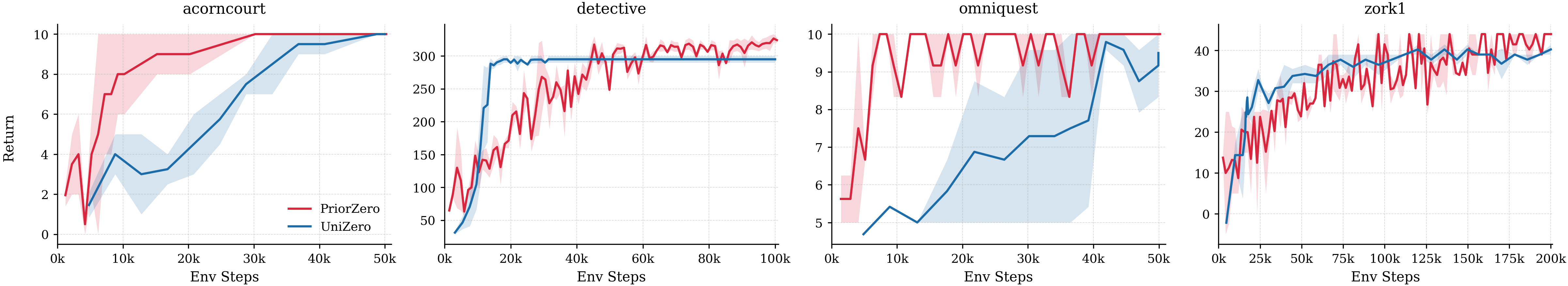}
    \caption{Main performance comparison between PriorZero and UniZero on four Jericho environments. Shaded regions indicate variation across three random seeds.}
    \vspace{-8pt}
    \label{fig:main_performance}
\end{figure}

Figure~\ref{fig:main_performance} compares PriorZero with UniZero on Jericho.
In \texttt{acorncourt} and \texttt{omniquest}, PriorZero reaches high-return regions substantially earlier than UniZero, indicating that semantic priors are useful when sparse rewards make classical RL exploration inefficient.
In \texttt{detective}, UniZero obtains strong early gains, but PriorZero continues improving and eventually surpasses it, suggesting that language priors become more valuable as they are adapted to task-specific dynamics.
In the long-horizon \texttt{zork1}, PriorZero achieves a higher asymptotic return, showing that root-prior guidance can still benefit planning even when success depends on extended multi-step trajectories.

Figure~\ref{fig:main_performance_llm} further examines the LLM component during alternating training. We refer to \textit{the standalone LLM policy} as PriorZero's LLM component evaluated in isolation: actions are sampled directly from $\pi_{\mathrm{LLM}}$ without any world-model rollout or MCTS, so the curve measures \emph{the LLM's own policy quality} as it is fine-tuned by the world-model advantage.
Across the four games, the standalone LLM policy improves steadily rather than collapsing under sparse rewards, confirming that the world-model-based advantage provides a stable, low-variance policy-gradient signal for LLM adaptation.
This trend explains why PriorZero's gains are not limited to the initial use of a frozen prior: as training proceeds, the LLM prior becomes increasingly aligned with environment dynamics, which improves root-node search guidance and in turn exposes the world model to higher-quality trajectories.

\vspace{-4pt}
\subsection{Analysis of Root-Prior Injection}
\vspace{-2pt}
\label{subsec:prior_analysis}

\begin{figure}[t]
    \centering
    \includegraphics[width=0.66\textwidth]{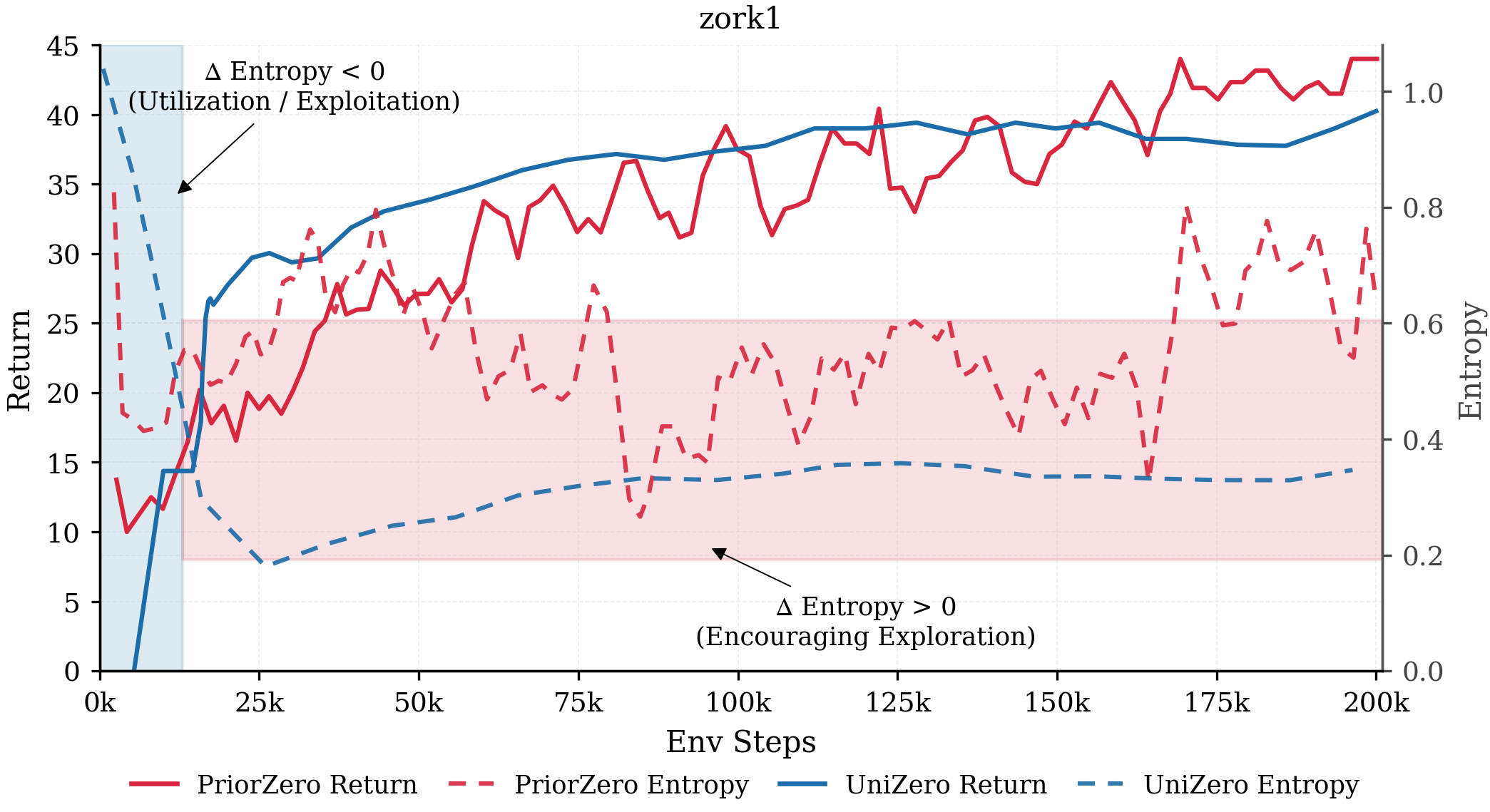}
    \caption{Return and MCTS root-node visit count entropy on \texttt{zork1} environment. Solid curves denote evaluation returns, and dashed curves denote MCTS root-node visit count entropy.}
    \label{fig:main_entropy}
    \vspace{-16pt}
\end{figure}

Figure~\ref{fig:main_entropy} tracks both return and MCTS root-node visit count entropy on \texttt{zork1}. Two phases emerge. Early in training, PriorZero improves faster than UniZero while its root-node entropy is \emph{lower}---the LLM prior concentrates search on semantically meaningful actions, avoiding cyclic exploration in \texttt{zork1}'s large admissible-action space. Later, as the world model matures, PriorZero's root entropy rises \emph{above} UniZero's while its return remains higher, showing that the prior does not collapse search onto a single action but instead prevents the visit distribution from being dominated by brittle world-model preferences. A representative case study and further breakdown are in Appendix~\ref{sec:appendix_priorzero_impl}.

Fig.~\ref{fig:frozen_prior_case} illustrates this effect at the action level.
In the game history, the agent repeatedly enters a closet to the north and returns south, forming an unproductive loop, while the current observation indicates that moving west leads back to the hallway.
The LLM assigns most probability mass to \texttt{west}, whereas the world model still favors \texttt{north}.
Root-prior fusion redirects MCTS toward the appropriate action and helps break the dead-loop exploration pattern.
More generally, these examples show that LLM priors can provide useful initial directions for MCTS, especially when the learned world model has not yet distinguished productive actions from locally repetitive but low-value behavior.

\vspace{-4pt}
\subsection{Ablation Studies}
\vspace{-2pt}
\label{subsec:ablation}

We conduct ablation studies on Jericho to validate the four key design choices of PriorZero:
(a)~the alternating fine-tuning schedule (\S\ref{subsec:decoupled_training}),
(b)~the chain-of-thought prior extraction with weighted token loss (\S\ref{subsec:root_injection}),
(c)~the capacity of the underlying LLM,
and (d)~the necessity of MCTS lookahead.

\begin{figure}[t]
    \centering
    \includegraphics[width=0.9\textwidth]{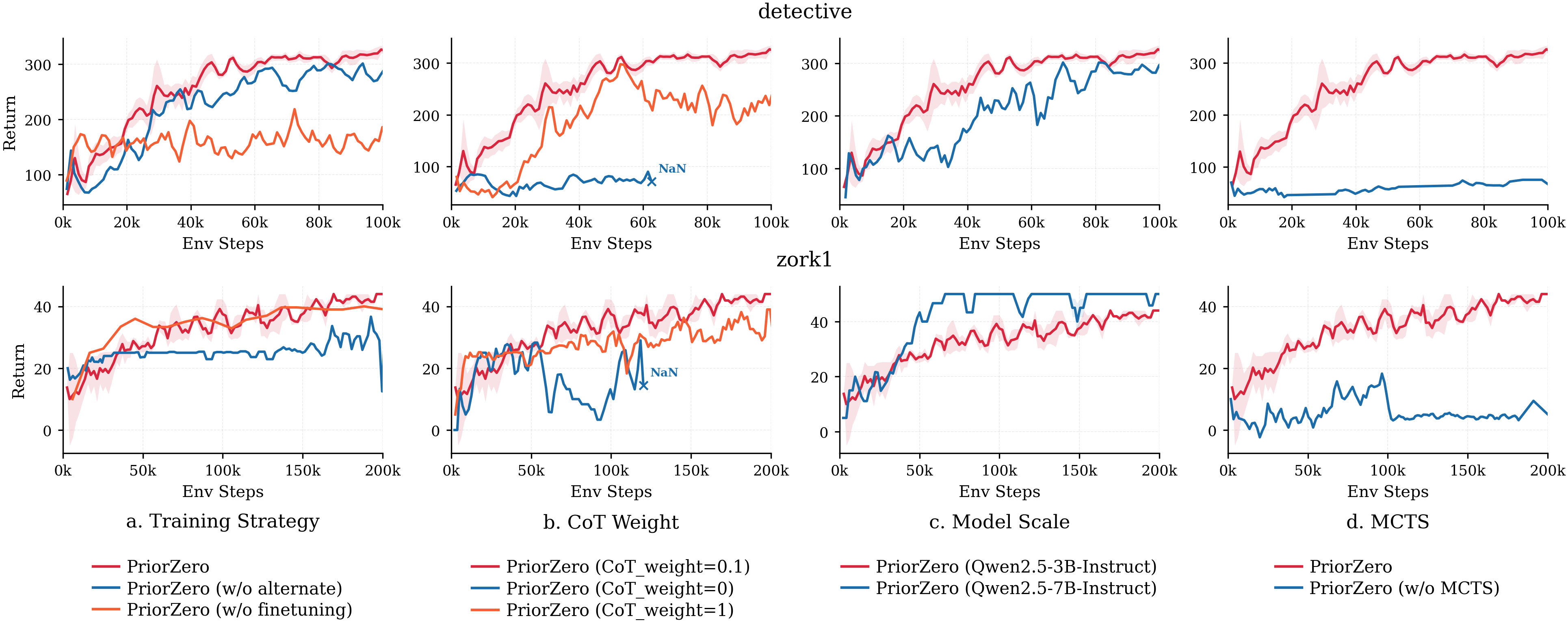}
    \vspace{-4pt}
    \caption{Ablation studies of the PriorZero on the \texttt{detective} and \texttt{zork1}. (a) Impact of the training strategy, comparing full PriorZero against a frozen-prior baseline and a non-alternating optimization variant. (b) Effect of Chain-of-Thought (CoT) reasoning and the corresponding CoT loss weight ($w_{cot}$). Note that removing CoT entirely leads to numerical instability (NaN). (c) Scaling the LLM prior from a 3B to a 7B parameter. (d) Contribution of MCTS planning to performance and stability.}
    \label{fig:ablation_all}
    \vspace{-5pt}
\end{figure}
\begin{figure}[t]
    \centering
    \begin{minipage}[c]{0.46\textwidth}
        \centering
        \includegraphics[width=\linewidth]{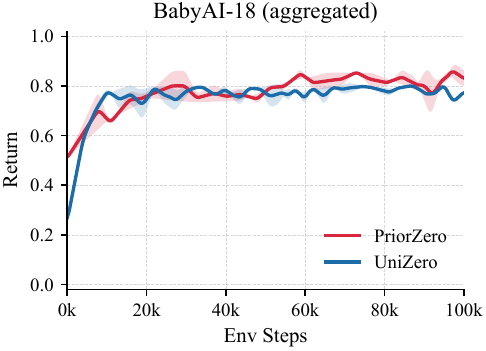}
    \end{minipage}\hfill
        \vspace{-12pt}
    \begin{minipage}[c]{0.42\textwidth}
        \caption{Aggregated return of PriorZero vs.\ UniZero on BabyAI-18, averaged over the 18 levels. Shaded regions indicate variation across three random seeds. PriorZero reaches high-return regions earlier and converges to a higher plateau, confirming that root-prior injection and alternating fine-tuning are also effective on the grid-world BabyAI benchmark.}
        \label{fig:babyai_train_reward}
    \end{minipage}
    \vspace{-12pt}
\end{figure}

\paragraph{Effect of LLM Fine-Tuning.}
Fig.~\ref{fig:ablation_all}a compares PriorZero against two variants: a frozen-LLM variant and a non-alternating variant.
While the frozen-LLM variant provides useful early-stage guidance—demonstrating the inherent strength of LLM priors, particularly in the challenging \texttt{zork1} environment—its performance eventually plateaus at a suboptimal level. 
In later training stages, the static LLM priors become increasingly misaligned with the agent's needs; because the LLM cannot be fine-tuned to adapt, these priors act as a bottleneck, leading to poor results even in the simpler \texttt{detective} game. Conversely, the non-alternating variant attempts to update the LLM from the very beginning. This approach, however, forces the LLM to learn from the purely noisy value signals of an uninitialized world model, leading to detrimental updates (or "knowledge corruption"). Consequently, this variant achieves only mediocre performance in \texttt{detective} and suffers a complete performance collapse in the difficult \texttt{zork1}. These findings underscore that PriorZero's efficacy stems not just from injecting LLM priors, but crucially from the asymmetric alternating schedule, which ensures the world model establishes reliable value estimates before they are used to guide the LLM's training.
\vspace{-10pt}
\paragraph{CoT Loss Weight Analysis.}
Figure~\ref{fig:ablation_all}b investigates the role of CoT reasoning and the loss weight $w_{\mathrm{cot}}$ applied to the reasoning-prefix tokens during LLM fine-tuning.
Completely omitting CoT reasoning induces severe training instability, resulting in numerical failures (NaN): without an intermediate reasoning prefix, the policy-gradient variance on raw action tokens becomes effectively unbounded under sparse rewards, producing NaN within a few iterations.
Conversely, applying a full CoT loss weight ($w_{\mathrm{cot}}{=}1.0$) prevents numerical collapse but leads to volatile learning dynamics and inferior final performance.
The likely cause is that strong supervision on the long reasoning prefix diffuses optimization capacity onto fine-grained reasoning-text tokens, while credit assignment within a long reasoning sequence is itself noisy---so heavy gradients on these tokens hurt rather than help.
Our default configuration ($w_{\mathrm{cot}}{=}0.1$) strikes a better balance: the action-token loss remains the dominant learning signal, while the reasoning prefix is preserved as a contextual scaffold.

\vspace{-3pt}
\paragraph{Scaling with LLM Size.}
Figure~\ref{fig:ablation_all}c assesses the scalability of LLM priors by upgrading the base model to \texttt{Qwen2.5-7B-Instruct}. In the long-horizon \texttt{zork1}, the 7B model exhibits faster convergence and achieves superior performance, confirming that more potent semantic priors significantly enhance exploration in expansive action spaces. Conversely, in the simpler \texttt{detective} task, while the 7B variant remains competitive, it suffers from higher variance compared to the 3B model, which converges more smoothly. This suggests that while scaling up the LLM bolsters semantic guidance—especially for complex tasks—it also introduces increased sensitivity to training dynamics, indicating that model scaling must be paired with robust optimization strategies to maintain stability.

\vspace{-3pt}
\paragraph{Effect of MCTS.}
Figure~\ref{fig:ablation_all}d isolates the impact of MCTS by disabling it, thereby forcing the agent to rely solely on the world model without lookahead planning. This variant collapses across both \texttt{detective} and \texttt{zork1}, and exhibits severe numerical instability in the former. This demonstrates that MCTS is an indispensable component of the PriorZero architecture: by fusing LLM semantic priors with world-model predictions, MCTS produces planning-improved policy targets that stabilize learning under sparse rewards---a stabilization role that no single component can play alone.
\vspace{-8pt}

\subsection{Extension to the Grid-World BabyAI Benchmark}
\label{subsec:babyai}

To test whether PriorZero benefits beyond text
games, we evaluate on the BabyAI~\citep{chevalier2019babyai}---a grid-world environment~\citep{xi2025agentgymrl}.
Agents navigate procedurally generated rooms, move objects, unlock doors, and interact with the environment in response to language commands.
Unlike Jericho's free-form text, BabyAI provides \emph{structured states} (object descriptions with relative positions), \emph{high-level semantic actions} (e.g., ``go to red ball 1''), and a \emph{continuous reward}.
Following~\citet{xi2025agentgymrl}, we train an agent across 18 levels spanning four categories.
For BabyAI we use a larger backbone, \texttt{Qwen2.5-7B-Instruct}, with an alternating schedule of $N_{\mathrm{WM}}{=}500$ world-model steps and $N_{\mathrm{LLM}}{=}100$ LLM steps ($5{:}1$ ratio); the increased LLM capacity matches the multi-task nature of BabyAI's 18 levels, while the smaller step counts reflect the shorter episodes.

Figure~\ref{fig:babyai_train_reward} reports aggregated return on the 18 BabyAI levels. PriorZero and UniZero learn at similar rates in the early phase, but in the later phase the LLM prior helps the world model escape local optima, lifting PriorZero to a higher asymptotic plateau ($0.82$ vs.\ $0.79$). The gap is most pronounced on compositional levels such as \texttt{SynthLoc}, where UniZero fails to learn any positive behavior while PriorZero attains $0.96$, indicating that semantic priors remain useful when agents must decompose goals into sub-tasks. Per-level curves, the standalone LLM ablation, and detailed setup are in Appendix~\ref{sec:appendix_implementation}.

\label{sec:related_work}
\vspace{-8pt}
\section{Related Work}
\vspace{-6pt}

\paragraph{Pre-Training for RL.}
Recent work integrates pre-training into RL to improve sample efficiency and generalization.
World Models~\citep{ha2018world} learn generative dynamics in latent space; CURL~\citep{laskin2020curl} applies contrastive learning in pixel space; PIE-G~\citep{yuan2022pre} transfers pre-trained encoders for zero-shot learning; VIP~\citep{ma2022vip} extracts implicit value from videos; and Dynalang~\citep{lin2024dynalang} trains world models on text corpora for planning without environments.
PriorZero does not require pre-training: the world model is trained online from scratch, while the LLM prior is adapted in-the-loop without relying on offline text datasets.
\vspace{-10pt}

\paragraph{LLM Priors for RL.}
LLMs have been leveraged to assist RL agents in multiple ways, which largely correspond to the paradigms categorized in \S\ref{sec:motivation}: as zero-shot planners~\citep{huang2022language} (P1), as end-to-end fine-tuned policies~\citep{carta2023grounding} (P2), as dynamics models~\citep{lin2024dynalang} (P3), as frozen text encoders~\citep{paischer2022history} (P4), and as MCTS action proposers~\citep{shi2025monte}.
The first four exhibit bottlenecks analyzed in \S\ref{sec:motivation}; the last keeps the LLM frozen and therefore cannot close the prior-dynamics mismatch through adaptation.
PriorZero is the first to integrate LLM priors with a learnable world model through root-prior injection and alternating fine-tuning, enabling continuous policy refinement while keeping dynamics learning uncontaminated by token-level bias.
\vspace{-10pt}
\paragraph{RL Fine-Tuning of LLMs.}
RLHF~\citep{christiano2017deep,ouyang2022traininglanguagemodelsfollow,xiong2026docr1} has become the dominant paradigm for LLM alignment, with extensions to AI feedback~\citep{bai2022constitutional,lee2023rlaif} and search-augmented training~\citep{guan2025rstar}. For decision-making, \citet{schmied2025llmsgreedyagentseffects} study RLFT effects on LLM decision behaviors but are limited to short-horizon tasks. PriorZero constructs advantage signals from world-model value estimates via value bootstrapping, providing low-variance feedback that resolves the credit-assignment failure limiting end-to-end RLFT (P2 in \S\ref{sec:motivation}) and enables stable fine-tuning in long-horizon, sparse-reward environments.
For a more comprehensive discussion of each line of work, please refer to Appendix~\ref{sec:appendix_related_work}.

\vspace{-8pt}
\section{Conclusion}
\vspace{-10pt}
\label{sec:conclusion}

We introduced PriorZero, a unified framework that addresses the \emph{prior-dynamics mismatch} between pretrained LLM and long-horizon RL optimization through a decoupled rollout-training design.
During rollout, root-only injection fuses LLM priors exclusively at the MCTS root node, guiding search toward semantically plausible actions while preserving the world model's multi-step lookahead capacity.
During training, alternating fine-tuning uses world-model value estimates to construct better advantage signals, providing stable policy-gradient for LLM adaptation.
Together, these components form a closed loop in which the world model and LLM mutually reinforce each other: the LLM's priors focus the search, and the world model's planning refines the LLM's priors.
Experiments on Jericho and BabyAI validate that PriorZero consistently improves exploration and return; ablations confirm that each component
is necessary, and that the framework scales with LLM capacity.

\begin{ack}
We extend our gratitude to several team-members of the Shanghai AI Laboratory for their invaluable assistance, support, and feedback on this paper and the associated codebase.
\end{ack}

\bibliographystyle{plainnat}
\bibliography{references}

@article{ha2018world,
  title={World models},
  author={Ha, David and Schmidhuber, J{\"u}rgen},
  journal={arXiv preprint arXiv:1803.10122},
  volume={2},
  number={3},
  pages={440},
  year={2018}
}

@inproceedings{laskin2020curl,
  title={Curl: Contrastive unsupervised representations for reinforcement learning},
  author={Laskin, Michael and Srinivas, Aravind and Abbeel, Pieter},
  booktitle={International conference on machine learning},
  pages={5639--5650},
  year={2020},
  organization={PMLR}
}

@inproceedings{seo2022reinforcement,
  title={Reinforcement learning with action-free pre-training from videos},
  author={Seo, Younggyo and Lee, Kimin and James, Stephen L and Abbeel, Pieter},
  booktitle={International Conference on Machine Learning},
  pages={19561--19579},
  year={2022},
  organization={PMLR}
}

@article{yuan2022pre,
  title={Pre-trained image encoder for generalizable visual reinforcement learning},
  author={Yuan, Zhecheng and Xue, Zhengrong and Yuan, Bo and Wang, Xueqian and Wu, Yi and Gao, Yang and Xu, Huazhe},
  journal={Advances in Neural Information Processing Systems},
  volume={35},
  pages={13022--13037},
  year={2022}
}

@article{ma2022vip,
  title={Vip: Towards universal visual reward and representation via value-implicit pre-training},
  author={Ma, Yecheng Jason and Sodhani, Shagun and Jayaraman, Dinesh and Bastani, Osbert and Kumar, Vikash and Zhang, Amy},
  journal={arXiv preprint arXiv:2210.00030},
  year={2022}
}

@article{wu2023pre,
  title={Pre-training contextualized world models with in-the-wild videos for reinforcement learning},
  author={Wu, Jialong and Ma, Haoyu and Deng, Chaoyi and Long, Mingsheng},
  journal={Advances in Neural Information Processing Systems},
  volume={36},
  pages={39719--39743},
  year={2023}
}

@inproceedings{basavatia2024starling,
  title={Starling: Self-supervised training of text-based reinforcement learning agent with large language models},
  author={Basavatia, Shreyas and Murugesan, Keerthiram and Ratnakar, Shivam},
  booktitle={Findings of the Association for Computational Linguistics: ACL 2024},
  pages={15804--15819},
  year={2024}
}

@article{li2022pre,
  title={Pre-trained language models for interactive decision-making},
  author={Li, Shuang and Puig, Xavier and Paxton, Chris and Du, Yilun and Wang, Clinton and Fan, Linxi and Chen, Tao and Huang, De-An and Aky{\"u}rek, Ekin and Anandkumar, Anima and others},
  journal={Advances in Neural Information Processing Systems},
  volume={35},
  pages={31199--31212},
  year={2022}
}

@article{ahn2022can,
  title={Do as i can, not as i say: Grounding language in robotic affordances},
  author={Ahn, Michael and Brohan, Anthony and Brown, Noah and Chebotar, Yevgen and Cortes, Omar and David, Byron and Finn, Chelsea and Fu, Chuyuan and Gopalakrishnan, Keerthana and Hausman, Karol and others},
  journal={arXiv preprint arXiv:2204.01691},
  year={2022}
}

@article{cao2024beyond,
  title={Beyond sparse rewards: Enhancing reinforcement learning with language model critique in text generation},
  author={Cao, Meng and Shu, Lei and Yu, Lei and Zhu, Yun and Wichers, Nevan and Liu, Yinxiao and Meng, Lei},
  journal={arXiv preprint arXiv:2401.07382},
  year={2024}
}

@article{liu2024dellma,
  title={Dellma: Decision making under uncertainty with large language models},
  author={Liu, Ollie and Fu, Deqing and Yogatama, Dani and Neiswanger, Willie},
  journal={arXiv preprint arXiv:2402.02392},
  year={2024}
}

@article{shi2025monte,
  title={Monte carlo planning with large language model for text-based game agents},
  author={Shi, Zijing and Fang, Meng and Chen, Ling},
  journal={arXiv preprint arXiv:2504.16855},
  year={2025}
}

@article{christiano2017deep,
  title={Deep reinforcement learning from human preferences},
  author={Christiano, Paul F and Leike, Jan and Brown, Tom and Martic, Miljan and Legg, Shane and Amodei, Dario},
  journal={Advances in neural information processing systems},
  volume={30},
  year={2017}
}

@article{ziegler2019fine,
  title={Fine-tuning language models from human preferences},
  author={Ziegler, Daniel M and Stiennon, Nisan and Wu, Jeffrey and Brown, Tom B and Radford, Alec and Amodei, Dario and Christiano, Paul and Irving, Geoffrey},
  journal={arXiv preprint arXiv:1909.08593},
  year={2019}
}

@article{stiennon2020learning,
  title={Learning to summarize with human feedback},
  author={Stiennon, Nisan and Ouyang, Long and Wu, Jeffrey and Ziegler, Daniel and Lowe, Ryan and Voss, Chelsea and Radford, Alec and Amodei, Dario and Christiano, Paul F},
  journal={Advances in neural information processing systems},
  volume={33},
  pages={3008--3021},
  year={2020}
}

@article{zhang2025stair,
  title={Stair: Improving safety alignment with introspective reasoning},
  author={Zhang, Yichi and Zhang, Siyuan and Huang, Yao and Xia, Zeyu and Fang, Zhengwei and Yang, Xiao and Duan, Ranjie and Yan, Dong and Dong, Yinpeng and Zhu, Jun},
  journal={arXiv preprint arXiv:2502.02384},
  year={2025}
}

@article{guan2025rstar,
  title={RStar-math: Small LLMs can master math reasoning with self-evolved deep thinking},
  author={Guan, Xinyu and Zhang, Li Lyna and Liu, Yifei and Shang, Ning and Sun, Youran and Zhu, Yi and Yang, Fan and Yang, Mao},
  journal={arXiv preprint arXiv:2501.04519},
  year={2025}
}

@inproceedings{xiong2026docr1,
  title={Docr1: Evidence page-guided grpo for multi-page document understanding},
  author={Xiong, Junyu and Wang, Yonghui and Zhao, Weichao and Liu, Chenyu and Yin, Bing and Zhou, Wengang and Li, Houqiang},
  booktitle={Proceedings of the AAAI Conference on Artificial Intelligence},
  volume={40},
  number={13},
  pages={11178--11186},
  year={2026}
}

@misc{schmied2025llmsgreedyagentseffects,
      title={LLMs are Greedy Agents: Effects of RL Fine-tuning on Decision-Making Abilities}, 
      author={Thomas Schmied and Jörg Bornschein and Jordi Grau-Moya and Markus Wulfmeier and Razvan Pascanu},
      year={2025},
      eprint={2504.16078},
      archivePrefix={arXiv},
      primaryClass={cs.LG},
      url={https://arxiv.org/abs/2504.16078}, 
}

@article{lee2023rlaif,
  title={Rlaif: Scaling reinforcement learning from human feedback with ai feedback},
  author={Lee, Harrison and Phatale, Samrat and Mansoor, Hassan and Lu, Kellie Ren and Mesnard, Thomas and Ferret, Johan and Bishop, Colton and Hall, Ethan and Carbune, Victor and Rastogi, Abhinav},
  year={2023}
}

@article{bai2022constitutional,
  title={Constitutional ai: Harmlessness from ai feedback},
  author={Bai, Yuntao and Kadavath, Saurav and Kundu, Sandipan and Askell, Amanda and Kernion, Jackson and Jones, Andy and Chen, Anna and Goldie, Anna and Mirhoseini, Azalia and McKinnon, Cameron and others},
  journal={arXiv preprint arXiv:2212.08073},
  year={2022}
}

@misc{ouyang2022traininglanguagemodelsfollow,
      title={Training language models to follow instructions with human feedback}, 
      author={Long Ouyang and Jeff Wu and Xu Jiang and Diogo Almeida and Carroll L. Wainwright and Pamela Mishkin and Chong Zhang and Sandhini Agarwal and Katarina Slama and Alex Ray and John Schulman and Jacob Hilton and Fraser Kelton and Luke Miller and Maddie Simens and Amanda Askell and Peter Welinder and Paul Christiano and Jan Leike and Ryan Lowe},
      year={2022},
      eprint={2203.02155},
      archivePrefix={arXiv},
      primaryClass={cs.CL},
      url={https://arxiv.org/abs/2203.02155}, 
}

@article{wang2023voyager,
  title={Voyager: An open-ended embodied agent with large language models},
  author={Wang, Guanzhi and Xie, Yuqi and Jiang, Yunfan and Mandlekar, Ajay and Xiao, Chaowei and Zhu, Yuke and Fan, Linxi and Anandkumar, Anima},
  journal={arXiv preprint arXiv:2305.16291},
  year={2023}
}

@inproceedings{chevalier2019babyai,
  title={BabyAI: A Platform to Study the Sample Efficiency of Grounded Language Learning},
  author={Chevalier-Boisvert, Maxime and Bahdanau, Dzmitry and Lahlou, Salem and Willems, Lucas and Saharia, Chitwan and Nguyen, Thien Huu and Bengio, Yoshua},
  booktitle={International Conference on Learning Representations},
  year={2019}
}

@inproceedings{xi2025agentgym,
  title={AgentGym: Evolving Large Language Model-based Agents across Diverse Environments},
  author={Xi, Zhiheng and Ding, Yiwen and Chen, Wenxiang and Hong, Boyang and Guo, Honglin and Wang, Junzhe and Guo, Xin and Yang, Dingwen and Liao, Chenyang and He, Wei and Gao, Songyang and Chen, Lu and Zheng, Rui and Zou, Yicheng and Gui, Tao and Zhang, Qi and Qiu, Xipeng and Huang, Xuanjing and Wu, Zuxuan and Jiang, Yu-Gang},
  booktitle={Proceedings of the 63rd Annual Meeting of the Association for Computational Linguistics},
  year={2025}
}

@article{xi2025agentgymrl,
  title={AgentGym-RL: Training LLM Agents for Long-Horizon Decision Making through Multi-Turn Reinforcement Learning},
  author={Xi, Zhiheng and Huang, Jixuan and Liao, Chenyang and Huang, Baodai and Guo, Honglin and Liu, Jiaqi and Zheng, Rui and Ye, Junjie and Zhang, Jiazheng and Chen, Wenxiang and He, Wei and Ding, Yiwen and Li, Guanyu and Chen, Zehui and Du, Zhengyin and Yao, Xuesong and Gui, Tao and Zhang, Qi and Qiu, Xipeng and Huang, Xuanjing},
  journal={arXiv preprint arXiv:2509.08755},
  year={2025}
}

@InProceedings{huang2022language,
  title = {Language Models as Zero-Shot Planners: Extracting Actionable Knowledge for Embodied Agents},
  author = {Huang, Wenlong and Abbeel, Pieter and Pathak, Deepak and Mordatch, Igor},
  booktitle = {Proceedings of the 39th International Conference on Machine Learning},
  pages = {9118--9147},
  year = {2022},
  volume = {162},
  series = {Proceedings of Machine Learning Research},
  publisher = {PMLR}
}

@InProceedings{huang2023inner,
  title = {Inner Monologue: Embodied Reasoning through Planning with Language Models},
  author = {Huang, Wenlong and Xia, Fei and Xiao, Ted and Chan, Harris and Liang, Jacky and Florence, Pete and Zeng, Andy and Tompson, Jonathan and Mordatch, Igor and Chebotar, Yevgen and Sermanet, Pierre and Jackson, Tomas and Brown, Noah and Luu, Linda and Levine, Sergey and Hausman, Karol and Ichter, Brian},
  booktitle = {Proceedings of The 6th Conference on Robot Learning},
  pages = {1769--1782},
  year = {2023},
  volume = {205},
  series = {Proceedings of Machine Learning Research},
  publisher = {PMLR}
}

@InProceedings{carta2023grounding,
  title = {Grounding Large Language Models in Interactive Environments with Online Reinforcement Learning},
  author = {Carta, Thomas and Romac, Cl{\'e}ment and Wolf, Thomas and Lamprier, Sylvain and Sigaud, Olivier and Oudeyer, Pierre-Yves},
  booktitle = {Proceedings of the 40th International Conference on Machine Learning},
  pages = {3676--3713},
  year = {2023},
  volume = {202},
  series = {Proceedings of Machine Learning Research},
  publisher = {PMLR}
}

@InProceedings{lin2024dynalang,
  title = {Learning to Model the World With Language},
  author = {Lin, Jessy and Du, Yuqing and Watkins, Olivia and Hafner, Danijar and Abbeel, Pieter and Klein, Dan and Dragan, Anca},
  booktitle = {Proceedings of the 41st International Conference on Machine Learning},
  pages = {29992--30017},
  year = {2024},
  volume = {235},
  series = {Proceedings of Machine Learning Research},
  publisher = {PMLR}
}

@InProceedings{paischer2022history,
  title = {History Compression via Language Models in Reinforcement Learning},
  author = {Paischer, Fabian and Adler, Thomas and Patil, Vihang and Bitto-Nemling, Angela and Holzleitner, Markus and Lehner, Sebastian and Eghbal-Zadeh, Hamid and Hochreiter, Sepp},
  booktitle = {Proceedings of the 39th International Conference on Machine Learning},
  pages = {17156--17185},
  year = {2022},
  volume = {162},
  series = {Proceedings of Machine Learning Research},
  publisher = {PMLR}
}

@InProceedings{hansen2024tdmpc2,
  title = {{TD-MPC2}: Scalable, Robust World Models for Continuous Control},
  author = {Hansen, Nick and Su, Hao and Wang, Xiaolong},
  booktitle = {Proceedings of the International Conference on Learning Representations},
  year = {2024}
}

@InProceedings{wang2024efficientzero,
  title = {EfficientZero V2: Mastering Discrete and Continuous Control with Limited Data},
  author = {Wang, Shengjie and Liu, Shaohuai and Ye, Weirui and You, Jiacheng and Gao, Yang},
  booktitle = {Proceedings of the 41st International Conference on Machine Learning},
  year = {2024}
}

@inproceedings{hausknecht2020jericho,
  title = {Interactive Fiction Games: A Colossal Adventure},
  author = {Hausknecht, Matthew and Ammanabrolu, Prithviraj and C{\^o}t{\'e}, Marc-Alexandre and Yuan, Xingdi},
  booktitle = {Proceedings of the AAAI Conference on Artificial Intelligence},
  year = {2020}
}

@inproceedings{kwon2023efficient,
  title = {Efficient Memory Management for Large Language Model Serving with PagedAttention},
  author = {Kwon, Woosuk and Li, Zhuohan and Zhuang, Siyuan and Sheng, Ying and Zheng, Lianmin and Yu, Cody Hao and Gonzalez, Joseph E and Zhang, Hao and Stoica, Ion},
  booktitle = {Proceedings of the 29th Symposium on Operating Systems Principles},
  pages = {611--626},
  year = {2023}
}

@inproceedings{szot2024llarp,
  title     = {Large Language Models as Generalizable Policies for Embodied Tasks},
  author    = {Szot, Andrew and Schwarzer, Max and Agrawal, Harsh and Mazoure, Bogdan and Talbott, Walter and Metcalf, Katherine and Mackraz, Natalie and Hjelm, Devon and Toshev, Alexander},
  booktitle = {International Conference on Learning Representations},
  year      = {2024}
}

@article{yu2026rwml,
  title   = {Reinforcement World Model Learning for LLM-based Agents},
  author  = {Yu, Xiao and Peng, Baolin and Xu, Ruize and Shen, Yelong and He, Pengcheng and Nath, Suman and Singh, Nikhil and Gao, Jiangfeng and Yu, Zhou},
  journal = {arXiv preprint arXiv:2602.05842},
  year    = {2026}
}

@inproceedings{schmied2023l2m,
  title     = {Learning to Modulate Pre-trained Models in RL},
  author    = {Schmied, Thomas and Hofmarcher, Markus and Paischer, Fabian and Pascanu, Razvan and Hochreiter, Sepp},
  booktitle = {Advances in Neural Information Processing Systems},
  year      = {2023}
}

@article{xuan2024rezero,
  title={Rezero: Boosting mcts-based algorithms by backward-view and entire-buffer reanalyze},
  author={Xuan, Chunyu and Niu, Yazhe and Pu, Yuan and Hu, Shuai and Liu, Yu and Yang, Jing},
  journal={arXiv preprint arXiv:2404.16364},
  year={2024}
}

@article{yao2022react,
  title={React: Synergizing reasoning and acting in language models},
  author={Yao, Shunyu and Zhao, Jeffrey and Yu, Dian and Du, Nan and Shafran, Izhak and Narasimhan, Karthik and Cao, Yuan},
  journal={arXiv preprint arXiv:2210.03629},
  year={2022}
}

@article{zhai2024fine,
  title={Fine-tuning large vision-language models as decision-making agents via reinforcement learning},
  author={Zhai, Yuexiang and Bai, Hao and Lin, Zipeng and Pan, Jiayi and Tong, Shengbang and Zhou, Yifei and Suhr, Alane and Xie, Saining and LeCun, Yann and Ma, Yi and others},
  journal={Advances in neural information processing systems},
  volume={37},
  pages={110935--110971},
  year={2024}
}

@inproceedings{hao2023reasoning,
  title={Reasoning with language model is planning with world model},
  author={Hao, Shibo and Gu, Yi and Ma, Haodi and Hong, Joshua and Wang, Zhen and Wang, Daisy and Hu, Zhiting},
  booktitle={Proceedings of the 2023 Conference on Empirical Methods in Natural Language Processing},
  pages={8154--8173},
  year={2023}
}

@inproceedings{xie2025making,
  title={Making large language models into world models with precondition and effect knowledge},
  author={Xie, Kaige and Yang, Ian and Gunerli, John and Riedl, Mark},
  booktitle={Proceedings of the 31st International Conference on Computational Linguistics},
  pages={7532--7545},
  year={2025}
}

@inproceedings{xu2025vlms,
  title={VLMs-Guided Representation Distillation for Efficient Vision-Based Reinforcement Learning},
  author={Xu, Haoran and Peng, Peixi and Tan, Guang and Chang, Yiqian and Li, Luntong and Tian, Yonghong},
  booktitle={Proceedings of the Computer Vision and Pattern Recognition Conference},
  pages={29534--29544},
  year={2025}
}

@article{wang2024llm,
  title={Llm-empowered state representation for reinforcement learning},
  author={Wang, Boyuan and Qu, Yun and Jiang, Yuhang and Shao, Jianzhun and Liu, Chang and Yang, Wenming and Ji, Xiangyang},
  journal={arXiv preprint arXiv:2407.13237},
  year={2024}
}

@inproceedings{yan2025efficient,
  title={Efficient reinforcement learning with large language model priors},
  author={Yan, Xue and Song, Yan and Feng, Xidong and Yang, Mengyue and Zhang, Haifeng and Ammar, Haitham Bou and Wang, Jun},
  booktitle={13th International Conference on Learning Representations Iclr 2025},
  pages={30818--30842},
  year={2025},
  organization={ICLR}
}

@article{niu2024lightzero,
  title={LightZero: A Unified Benchmark for Monte Carlo Tree Search in General Sequential Decision Scenarios},
  author={Niu, Yazhe and Pu, Yuan and Yang, Zhenjie and Li, Xueyan and Zhou, Tong and Ren, Jiyuan and Hu, Shuai and Li, Hongsheng and Liu, Yu},
  journal={Advances in Neural Information Processing Systems},
  volume={36},
  year={2024}
}

@article{unizero,
  title={UniZero: Generalized and Efficient Planning with Scalable Latent World Models},
  author={Pu, Yuan and Niu, Yazhe and Yang, Zhenjie and Ren, Jiyuan and Li, Hongsheng and Liu, Yu},
  journal={Transactions on Machine Learning Research}
}

@article{pu2025one,
  title={One Model for All Tasks: Leveraging Efficient World Models in Multi-Task Planning},
  author={Pu, Yuan and Niu, Yazhe and Tang, Jia and Xiong, Junyu and Hu, Shuai and Li, Hongsheng},
  journal={arXiv preprint arXiv:2509.07945},
  year={2025}
}

@misc{lightrft,
  title={LightRFT: Light, Efficient, Omni-modal \& Reward-model Driven Reinforcement Fine-Tuning Framework},
  author={Niu, Yazhe and Pu, Yuan and Shi, Dongxing and Lu, Yudong and Xiong, Yingtong and Ge, Ruijun and Sun, Jiaxuan and Wan, Zunian and Zhang, Shaoang},
  publisher={GitHub},
  howpublished={\url{https://github.com/opendilab/LightRFT}},
  year={2025},
}

@article{hu2024openrlhf,
  title={OpenRLHF: An Easy-to-use, Scalable and High-performance RLHF Framework},
  author={Jian Hu and Xibin Wu and Zilin Zhu and Xianyu and Weixun Wang and Dehao Zhang and Yu Cao},
  journal={arXiv preprint arXiv:2405.11143},
  year={2024}
}

\appendix

\clearpage
\section*{Appendix Contents}
\phantomsection
\addcontentsline{toc}{section}{Appendix Contents}

\begin{itemize}
    \item[\ref{sec:appendix_paradigms}.] \hyperlink{sec:appendix_paradigms}{\nameref{sec:appendix_paradigms}} \dotfill \pageref{sec:appendix_paradigms}
    \item[\ref{sec:appendix_implementation}.] \hyperlink{sec:appendix_implementation}{\nameref{sec:appendix_implementation}} \dotfill \pageref{sec:appendix_implementation}
        \item[\ref{sec:appendix_related_work}.] \hyperlink{sec:appendix_related_work}{\nameref{sec:appendix_related_work}} \dotfill \pageref{sec:appendix_related_work}
    \item[\ref{sec:appendix_prompt}.] \hyperlink{sec:appendix_prompt}{\nameref{sec:appendix_prompt}} \dotfill \pageref{sec:appendix_prompt}
\end{itemize}
\clearpage

\section{Detailed Analysis of LLM Knowledge Transfer Paradigms}
\label{sec:appendix_paradigms}
\begin{figure*}[hb]
    \centering
        \includegraphics[width=\linewidth]{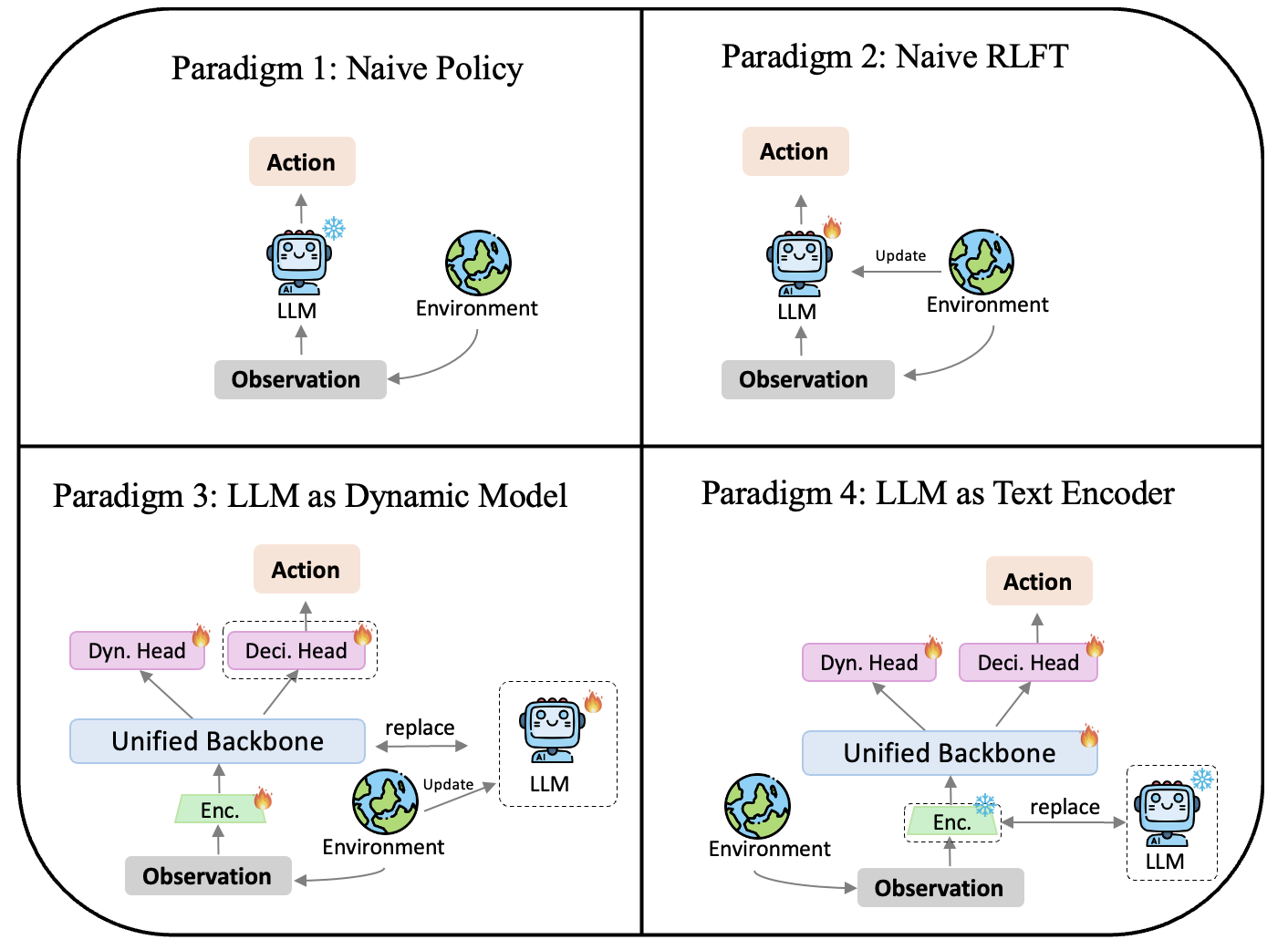}
    \caption{Comparison of the four LLM-knowledge-transfer paradigms studied in our ablations. Each panel shows where the LLM is inserted into the RL pipeline (as the policy, as a trainable policy, as the dynamics model backbone, or as the text encoder), with an icon in the top-right corner of each module indicating its training status: a fire icon denotes trainable components, while a ice icon denotes fixed (frozen) components.}
    \label{fig:paradigms_comparison}
\end{figure*}

We provide detailed formulations and analysis of the four existing paradigms for transferring LLM knowledge to RL agents, summarized in Table~\ref{tab:paradigms_comparison} and illustrated in Figure~\ref{fig:paradigms_comparison}. These paradigms correspond to four natural roles an LLM may assume in a text-based RL pipeline: as the policy, as a trainable policy, as the dynamics model backbone, or as the text encoder, as visualized in the four panels of Figure~\ref{fig:paradigms_comparison}.
Unless otherwise stated, all ablations use the same Jericho task setup and admissible action interface as the main PriorZero experiments.

\paragraph{Paradigm 1: LLM Directly as Policy (Naive Policy).}
The most intuitive approach is to use the LLM itself as a zero-shot or few-shot policy~\citep{huang2022language,ahn2022can,wang2023voyager}. At each environment step, the current textual observation, a bounded interaction history, and the valid action set returned by Jericho are converted into a natural-language prompt. In our implementation, the history length is fixed to $25$ previous transitions, and the prompt explicitly lists the current valid actions. A frozen Qwen2.5-3B-Insrtuct model then scores each admissible action string. Formally, for an action $a\in\mathcal{A}_t$ with tokenization $y(a)$, we compute
\begin{equation}
    s_\theta(a \mid o_t,h_t,\mathcal{A}_t)
    =
    \frac{1}{|y(a)|}
    \sum_{i=1}^{|y(a)|}
    \log p_\theta\!\left(
        y_i(a)
        \mid
        \mathrm{Prompt}(o_t,h_t,\mathcal{A}_t), y_{<i}(a)
    \right),
\end{equation}
where $o_t$ is the current observation, $h_t$ is the truncated history, and $\mathcal{A}_t$ is the valid action set. The induced policy is obtained by temperature-normalizing the action scores,
\begin{equation}
    \pi_{\mathrm{LLM}}(a\mid o_t,h_t)
    =
    \frac{\exp(s_\theta(a\mid o_t,h_t,\mathcal{A}_t)/\tau)}
    {\sum_{a'\in\mathcal{A}_t}\exp(s_\theta(a'\mid o_t,h_t,\mathcal{A}_t)/\tau)}.
\end{equation}
The greedy variant (used in this paper) executes $\arg\max_{a\in\mathcal{A}_t}\pi_{\mathrm{LLM}}(a\mid o_t,h_t)$, while the stochastic variant samples from this distribution.
This baseline deliberately removes the world model, MCTS, replay buffer, value learning, and all forms of training. It therefore isolates the effect of the pretrained LLM prior when used as a complete decision rule. Although this paradigm is simple and can exploit commonsense affordances encoded in the LLM, it is fundamentally limited by the \emph{knowing-doing gap}: high language-model likelihood does not imply high environment return. The model may prefer actions that are linguistically plausible or locally sensible, but it has no mechanism for environment-specific adaptation---no temporal credit assignment, no belief-state maintenance, no systematic exploration, and no multi-step lookahead. Consequently, errors made early in an episode cannot be corrected through policy improvement, and the agent stagnates in long-horizon settings.

\paragraph{Paradigm 2: End-to-End RL Fine-Tuning (Naive RLFT).}
The second paradigm closes the environment-feedback loop by treating the LLM as a parameterized policy and fine-tuning it with reinforcement learning. We instantiate this baseline as a standard actor-critic RLFT procedure. The actor is initialized from the LLM policy, and a scalar value head is attached to the same backbone. A frozen reference model is retained to regularize updates through a KL penalty. At iteration $k$, the current policy collects $N$ complete Jericho episodes, all step-level samples are merged into a rollout buffer, and the value head is used to compute generalized advantage estimates. Specifically,
\begin{equation}
    \delta_t = r_t + \gamma V_\phi(s_{t+1}) - V_\phi(s_t),
    \qquad
    \hat{A}_t = \sum_{\ell=0}^{T-t}(\gamma\lambda)^\ell \delta_{t+\ell},
\end{equation}
with return target $\hat{R}_t=\hat{A}_t+V_\phi(s_t)$. The implementation uses the value predicted at the prompt boundary, i.e., the state representation immediately before the action tokens, so that the value corresponds to the current game state rather than to the generated action text.

Policy optimization then follows the clipped PPO objective:
\begin{equation}
    \mathcal{L}_{\mathrm{PPO}}
    =
    -\mathbb{E}_{t}
    \left[
    \min\left(
        \rho_t \hat{A}_t,
        \mathrm{clip}(\rho_t,1-\epsilon,1+\epsilon)\hat{A}_t
    \right)
    \right]
    + c_v \mathcal{L}_{V}
    + \beta\,\mathrm{KL}\!\left(\pi_\theta\,\Vert\,\pi_{\mathrm{ref}}\right),
\end{equation}
where $\rho_t=\pi_\theta(a_t\mid s_t)/\pi_{\theta_{\mathrm{old}}}(a_t\mid s_t)$, $\mathcal{L}_{V}$ is a clipped value regression loss, and $\pi_{\mathrm{ref}}$ is the frozen reference policy. In our ablation, one training iteration collects a fixed number of full episodes (default $50$), normalizes advantages across all the workers, and performs multiple PPO epochs over minibatches sampled from the merged rollout data. The full experiment is budgeted by total environment steps rather than by gradient steps, matching the reporting convention used by PriorZero.

This paradigm is appealing because it directly optimizes the LLM for task reward and allows the policy to adapt beyond static prompting. However, in Jericho-style environments it faces a difficult combination of sparse rewards, long horizons, and large language-action spaces. The LLM must simultaneously learn action grounding, exploration, value estimation, and long-horizon credit assignment from expensive on-policy rollouts. Even with PPO clipping and reference KL regularization, useful gradients are delayed and high variance, while each update requires forward/backward passes through a large language model. Thus RLFT can improve local action preferences but remains computationally expensive and statistically inefficient as a standalone solution.
  
\paragraph{Paradigm 3: LLM as Dynamics Model (Latent World-Model Backbone).}
A third paradigm retains the model-based RL structure but replaces the latent dynamics backbone with an LLM.
In this ablation, we keep the UniZero-style interaction loop, replay buffer, MCTS, reward/value/policy heads, and the lightweight BGE observation encoder. The only architectural substitution is inside the latent world model: the transformer that processes interleaved latent observation tokens and action tokens is replaced by Qwen2.5-0.5B. Thus the LLM is not used to read raw observations or directly generate actions; it serves as the sequence backbone for latent transition modeling.

Let $z_t=f_{\mathrm{BGE}}(o_t)$ denote the observation embedding and $e(a_t)$ denote the learned action embedding. The world model receives sequences of the form
\begin{equation}
    x_{1:2K} = (z_{t-K+1}, e(a_{t-K+1}), \ldots, z_t, e(a_t)),
\end{equation}
and predicts next-latent observations, rewards, values, and policies from the Qwen hidden states. Because Qwen2.5 uses grouped-query attention, the implementation adapts the UniZero KV cache so that the number of key/value heads and the per-head hidden dimension match the pretrained Qwen configuration. The downstream UniZero heads are trained with the same world-model losses as the standard backbone, including latent reconstruction, reward, policy and value prediction.

This ablation tests whether next-token pretraining provides a useful inductive bias for latent dynamics. The limitation is a \emph{granularity mismatch}: a causal LLM is pretrained at the token level to model natural-language continuations, while a latent world model must operate at the state-action level to represent action-conditioned, controllable state transitions. These two structures are related but not equivalent---linguistically similar states may have different transition consequences, and dynamically equivalent states may be described with different text. As a result, the world knowledge encoded in token-level pretraining cannot be effectively reused for latent dynamics learning: replacing the dynamics backbone with a pretrained LLM imports semantic structure but also imposes a representation geometry poorly aligned with one-step transition accuracy and multi-step MCTS rollout consistency. This paradigm therefore probes whether LLM capacity can substitute for a task-trained dynamics model; our results indicate that it cannot.

\paragraph{Paradigm 4: LLM as Text Encoder.}
The fourth paradigm places the LLM at the perception layer. Instead of using a lightweight BGE encoder to convert observations into latent vectors, this ablation uses Qwen3-4B as the text encoder. Given an observation token sequence $x_{1:L}$, the causal LLM produces final-layer hidden states $h_{1:L}$. Since the hidden state at the last non-padding token is the representation used to predict the next token, we use that vector as the sentence-level embedding and project it into the UniZero latent dimension:
\begin{equation}
    z_t
    =
    \mathrm{Norm}\!\left(W h_L\right),
\end{equation}
where $W$ is a learnable projection head and $\mathrm{Norm}$ is the same encoder normalization used by the world model. The pretrained Qwen encoder is frozen, while the projection and downstream RL/world-model modules are optimized by the usual training objective.

This paradigm avoids catastrophic forgetting and provides a stronger semantic encoder than a lightweight sentence model. Nevertheless, it introduces a \emph{representation bottleneck}: the frozen next-token representation is optimized to support language continuation, not to preserve the controllable variables required by planning. Textual observations that are close under semantic similarity can require different actions, whereas observations with different surface forms can correspond to the same underlying game state. Because the LLM encoder is frozen, the RL agent can only adapt the projection and downstream dynamics model, limiting its ability to reshape the representation around task-specific invariances and action-conditioned distinctions. The resulting non-adaptive representation imposes a performance ceiling---the agent benefits from richer language features but cannot surpass the expressiveness boundary set by the frozen encoder.

\section{Implementation and Experiment Details}
\label{sec:appendix_implementation}
\label{sec:appendix_exp}


\subsection{PriorZero Implementation Details}
\label{sec:appendix_priorzero_impl}

\noindent Our code is available at \textcolor{magenta}{\url{https://github.com/opendilab/LightZero}}.

This subsection gives the complete implementation-level view of PriorZero.
The actual system contains three coupled but gradient-decoupled components: (i) a UniZero-style world model with latent MCTS, (ii) an LLM prior module that scores admissible action strings through prompt log-probabilities, and (iii) an alternating trainer that uses world-model value estimates to fine-tune the LLM, with the LLM rollout-and-update infrastructure (vLLM-based generation, PPO token loss, KL regularization, and reference-model bookkeeping) adapted from LightRFT~\citep{lightrft} and OpenRLHF~\citep{hu2024openrlhf}.
The two trainable modules communicate only through rollout data and scalar training targets: LLM priors affect the MCTS root distribution and thereby the collected trajectories, while world-model values produce the advantage signal used to update the LLM.
Table~\ref{tab:priorzero_pseudocode} summarizes the overall PriorZero training procedure in pseudocode form.

\begin{table}[t]
\centering
\caption{Pseudocode of PriorZero. The LLM prior is injected only at the MCTS root during rollout, while LLM fine-tuning is driven by  value-based advantage computed from the target world model.}
\label{tab:priorzero_pseudocode}
\small
\setlength{\tabcolsep}{4pt}
\renewcommand{\arraystretch}{1.08}
\begin{tabularx}{\textwidth}{@{}rL@{}}
\toprule
\multicolumn{2}{@{}>{\raggedright\arraybackslash}p{\textwidth}@{}}{\textbf{Input:} environment $\mathcal{E}$, admissible actions $\mathcal{A}(s)$, MCTS budget $K$, TD horizon $n$, fusion weight $\alpha$, schedules $N_{\mathrm{WM}},N_{\mathrm{LLM}}$} \\
\multicolumn{2}{@{}>{\raggedright\arraybackslash}p{\textwidth}@{}}{\textbf{Initialize:} world model $\theta$, target world model $\bar{\theta}\leftarrow\theta$, LLM $\phi$, reference LLM $\phi_{\mathrm{ref}}$, replay buffer $\mathcal{D}$, phase $p\leftarrow\mathrm{WM}$} \\
\midrule
1  & \textbf{while} training budget is not exhausted \textbf{do} \\
2  & \quad \textbf{if} $p=\mathrm{WM}$ or collection is enabled in the LLM phase \textbf{then} \\
3  & \quad\quad \textbf{for} each environment step $t$ \textbf{do} \\
4  & \quad\quad\quad $\mathcal{C}_t \leftarrow \operatorname{BuildPrompt}(o_t,\mathcal{H}_t,\mathcal{A}(s_t))$ \\
5  & \quad\quad\quad $(\pi_{\mathrm{LLM}}^T,\mathcal{B}_t) \leftarrow \operatorname{ScoreActionsWithLLM}(\phi,\mathcal{C}_t,\mathcal{A}(s_t))$ \\
6  & \quad\quad\quad $(z_t,\pi_{\mathrm{WM}},v_{\theta}) \leftarrow \operatorname{InitialInference}_{\theta}(s_t)$ \\
7  & \quad\quad\quad $P_{\mathrm{root}}\leftarrow(1-\alpha)\pi_{\mathrm{WM}}+\alpha\pi_{\mathrm{LLM}}^T$ \hfill \textit{root only} \\
8  & \quad\quad\quad $(\pi_{\mathrm{MCTS}},V_t)\leftarrow\operatorname{MCTS}(z_t,P_{\mathrm{root}},K;\theta)$ \hfill \textit{internal nodes use $\pi_{\mathrm{WM}}$} \\
9  & \quad\quad\quad $a_t\sim\pi_{\mathrm{MCTS}}(\cdot\mid s_t)$; \quad $(o_{t+1},r_t,\mathrm{done})\leftarrow\mathcal{E}.\operatorname{Step}(a_t)$ \\
10 & \quad\quad\quad $\mathcal{D}\leftarrow\mathcal{D}\cup(o_t,\mathcal{H}_t,a_t,r_t,\pi_{\mathrm{MCTS}},V_t,\mathcal{B}_t)$ \\
11 & \quad\quad \textbf{end for} \\
12 & \quad \textbf{end if} \\
13 & \quad \textbf{if} $p=\mathrm{WM}$ \textbf{then} \\
14 & \quad\quad \textbf{for} $i=1,\ldots,N_{\mathrm{WM}}$ \textbf{do} \\
15 & \quad\quad\quad $\mathcal{M}\leftarrow\operatorname{SampleWindows}(\mathcal{D})$; update $\theta$ with reward, value, policy, and dynamics losses \\
16 & \quad\quad\quad $\bar{\theta}\leftarrow\operatorname{UpdateTarget}(\bar{\theta},\theta)$ \\
17 & \quad\quad \textbf{end for}; \quad $p\leftarrow\mathrm{LLM}$ \\
18 & \quad \textbf{else} \\
19 & \quad\quad \textbf{for} $i=1,\ldots,N_{\mathrm{LLM}}$ \textbf{do} \\
20 & \quad\quad\quad $\mathcal{M}\leftarrow\operatorname{FetchLLMSamples}(\mathcal{D})$ \\
21 & \quad\quad\quad $Q_t^{(n)}\leftarrow\sum_{j=0}^{n-1}\gamma^j r_{t+j}+\gamma^n v_{\bar{\theta}}(s_{t+n})$ \\
22 & \quad\quad\quad $A_t\leftarrow Q_t^{(n)}-v_{\bar{\theta}}(s_t)$; \quad $\hat{A}\leftarrow\operatorname{Normalize}(A,r_{\mathrm{fmt}})$ \\
23 & \quad\quad\quad update $\phi$ with clipped PPO token loss and KL to $\phi_{\mathrm{ref}}$ \\
24 & \quad\quad\quad synchronize $\phi$ to the vLLM rollout engine \\
25 & \quad\quad \textbf{end for}; \quad $p\leftarrow\mathrm{WM}$ \\
26 & \quad \textbf{end if} \\
27 & \textbf{end while} \\
\bottomrule
\end{tabularx}
\end{table}

\paragraph{Planning advantage and comparison with GRPO.}
For LLM fine-tuning, PriorZero does not use a group-relative reward baseline.
For each sampled transition, the replay buffer computes two world-model value quantities using the target model $\bar{\theta}$: a bootstrapped TD target from the future state and a predicted value at the current state.
The raw advantage is
\[
A_t
=
Q^{(n)}_{\bar{\theta}}(s_t,a_t)-v_{\bar{\theta}}(s_t),
\qquad
Q^{(n)}_{\bar{\theta}}(s_t,a_t)
=
\sum_{i=0}^{n-1}\gamma^i r_{t+i}
+\gamma^n v_{\bar{\theta}}(s_{t+n}).
\]
In the default configuration, these raw advantages are accumulated within the current LLM phase and normalized by global phase statistics,
\[
\hat{A}_t=\frac{A_t-\mu_{\mathrm{phase}}}{\sigma_{\mathrm{phase}}+\epsilon}.
\]

When CoT format supervision is enabled, the final scalar used by the policy loss is
\[
\hat{A}^{\mathrm{final}}_t
=
(1-\lambda_{\mathrm{fmt}})\hat{A}_t
+
\lambda_{\mathrm{fmt}}r_{\mathrm{fmt}},
\]
where $r_{\mathrm{fmt}}\in\{0,1\}$ checks whether the LLM output follows the required \texttt{Reasoning: ... Action: ...} template.

This differs from GRPO-style training in the source of the baseline and in the unit of comparison.
GRPO samples multiple completions for the same prompt and constructs a relative advantage by subtracting the mean reward of that prompt's completion group, often also dividing by the group standard deviation:
\[
A^{\mathrm{GRPO}}_{i}
\approx
\frac{R_i-\frac{1}{G}\sum_{j=1}^{G}R_j}
{\mathrm{std}(\{R_j\}_{j=1}^{G})+\epsilon}.
\]
Such a baseline avoids learning an explicit value function, but it requires several rollouts per prompt and its signal is only relative within the sampled group.
In sparse, long-horizon environments, most group rewards can be identical or delayed, making the normalized reward signal noisy or uninformative.
PriorZero instead uses the world model as a state-dependent critic: it combines real short-horizon rewards with bootstrapped long-term value and compares the resulting target against the same state's baseline.
This gives lower-variance credit assignment without requiring multiple LLM completions for every prompt, while preserving an absolute notion of whether the executed action improved over the current value estimate.

This estimator is \emph{approximately} off-policy: actions $a_t$ are drawn from
the MCTS visit distribution $\pi_{\mathrm{MCTS}}$ rather than from $\pi_\phi$
directly, so $v_{\bar\theta}$ acts as a proxy for $V^{\pi_\phi}$ rather than an
exact on-policy value. Two factors keep this approximation tight in our setting.
First, the LLM prior is injected at the MCTS root (line~7 of
Table~\ref{tab:priorzero_pseudocode}), so $\pi_\phi$ and $\pi_{\mathrm{MCTS}}$
are coupled by construction: the search distribution is a planning-improved
version of the LLM prior on the same root state, not an independently sampled
behaviour policy. Second, the clipped PPO ratio
$\rho_{t,j}\in[1-\epsilon_{\mathrm{low}},1+\epsilon_{\mathrm{high}}]$ together
with the KL regularizer $\beta D_{\mathrm{KL}}(\pi_\phi\Vert\pi_{\phi_{\mathrm{ref}}})$
bounds the per-update policy drift, which limits how far $\pi_\phi$ can move
away from the behaviour distribution that produced the stored advantages,
making this a standard near-on-policy regime in PPO-style fine-tuning. Such
crucial under Jericho-style sparse rewards, where any additional variance in
the policy-gradient signal would be amplified by long credit-assignment chains.

The trade-off is that PriorZero depends on the world model's value quality,
which is why the implementation uses a WM warm-up and asymmetric alternating
updates with many more WM updates than LLM updates.

\paragraph{PriorZero loss vs. AlphaZero-like SFT loss.}
An AlphaZero-like supervised objective for the LLM prior would train the LLM to imitate the search-improved policy:
\[
\mathcal{L}_{\mathrm{AZ\text{-}SFT}}(\phi)
=
-
\mathbb{E}_{s_t\sim\mathcal{D}}
\sum_{a\in\mathcal{A}_t}
\pi_{\mathrm{MCTS}}(a\mid s_t)\log \pi_{\phi}(a\mid \mathcal{C}_t).
\]
For a language model implementation, this is usually approximated by SFT on the selected action string, possibly weighted by the MCTS visit count:
\[
\mathcal{L}_{\mathrm{SFT}}(\phi)
=
-
w_t\sum_{j\in y(a_t)}\log \pi_{\phi}(y_j(a_t)\mid \mathcal{C}_t,y_{<j}(a_t)).
\]
This objective is simple, but it treats the planner output as a label to imitate.
It does not directly encode whether the chosen action is better or worse than the current value baseline, and the update direction is always to increase the likelihood of the supervised target.
As a result, a noisy MCTS target early in training can be copied into the LLM prior, and the loss has no clipped policy-improvement mechanism to control update size.

PriorZero instead optimizes a clipped policy-gradient objective over the stored action tokens:
\[
\rho_{t,j}(\phi)
=
\exp\!\left(
\log \pi_{\phi}(y_j\mid \mathcal{C}_t,y_{<j})
-
\log \pi_{\phi_{\mathrm{old}}}(y_j\mid \mathcal{C}_t,y_{<j})
\right),
\]
\[
\mathcal{L}_{\mathrm{PZ}}(\phi)
=
-
\mathbb{E}_{t,j}
\left[
m_{t,j}\min\!\left(
\rho_{t,j}\hat{A}^{\mathrm{final}}_t,\;
\mathrm{clip}(\rho_{t,j},1-\epsilon_{\mathrm{low}},1+\epsilon_{\mathrm{high}})
\hat{A}^{\mathrm{final}}_t
\right)
\right]
+
\beta D_{\mathrm{KL}}(\pi_{\phi}\Vert\pi_{\phi_{\mathrm{ref}}}).
\]
Here $m_{t,j}$ is the action-token mask; when CoT is used, reasoning-prefix tokens receive a smaller weight $w_{\mathrm{cot}}$ so that the RL signal focuses on the final action while still preserving the structured reasoning format.
Positive advantages increase the probability of actions whose TD target exceeds the baseline, while negative advantages decrease actions that underperform the baseline.
Thus the LLM is not merely distilled from MCTS; it is updated according to whether the action actually improved the world-model value estimate under environment feedback.

\paragraph{Fusion of the LLM prior and world-model policy.}
The implementation fuses the two policies in probability space rather than adding raw logits.
This is important because LLM prompt log-probabilities and world-model policy logits have different scales.
Concretely, invalid actions are first masked out, the LLM scores are temperature-normalized over the admissible action set, and the world-model logits are separately converted to $\pi_{\mathrm{WM}}$.
The fused root logits are then obtained by taking the logarithm of a convex mixture:
\[
\ell_{\mathrm{root}}(a)
=
\log\!\left(
\lambda_{\mathrm{WM}}\pi_{\mathrm{WM}}(a\mid z_t)
+
(1-\lambda_{\mathrm{WM}})\pi_{\mathrm{LLM}}^T(a\mid\mathcal{C}_t)
+\epsilon
\right).
\]
The default experiments use a fixed mixture with $\lambda_{\mathrm{WM}}=0.5$.
The code also supports an adaptive variant that sets the LLM coefficient from the normalized entropy of the world-model root policy:
\[
\alpha_t
=
\alpha_{\min}
+
(\alpha_{\max}-\alpha_{\min})
\left(
1-\frac{H(\pi_{\mathrm{WM}}(\cdot\mid z_t))}{\log |\mathcal{A}_t|}
\right),
\qquad
P_{\mathrm{root}}
=(1-\alpha_t)\pi_{\mathrm{WM}}+\alpha_t\pi_{\mathrm{LLM}}^T.
\]
In all modes, the fusion is applied only at the root node.
This design reflects three implementation considerations.
First, the root state corresponds to a real textual observation, so the LLM prior is semantically grounded there; deeper MCTS nodes are imagined latent states and may not have faithful textual descriptions.
Second, root-only fusion prevents the LLM from overwriting the world model's action-conditioned dynamics during multi-step lookahead.
Third, the fused root distribution affects the world model indirectly through data and MCTS visit-count targets, while gradients remain separated: the world model learns from collected trajectories and search targets, and the LLM learns from value advantages computed after replay sampling.
This keeps the LLM-prior and WM-policy roles complementary: the LLM proposes where search should begin, and the world model decides which branches are valuable through latent planning.


\subsection{Jericho Details}
\label{sec:appendix_jericho}

\subsubsection{Environment Details}

We evaluate on four Jericho text adventure games spanning a range of difficulty levels. Table~\ref{tab:jericho_games} lists the games with their action space sizes and maximum episode lengths as configured in the codebase.

\begin{table}[htbp]
\centering
\caption{Jericho games information, with per-game action space size and maximum episode steps.}
\label{tab:jericho_games}
\small
\begin{tabular}{lccc}
\toprule
\textbf{Game} & \textbf{Game File} & \textbf{Action Space Size} & \textbf{Max Steps} \\
\midrule
Detective   & \texttt{detective.z5}   & 12  & 100 \\
OmniQuest   & \texttt{omniquest.z5}   & 25  & 100 \\
Acorn Court & \texttt{acorncourt.z5}  & 45  & 50  \\
Zork I      & \texttt{zork1.z5}       & 55  & 500 \\
\bottomrule
\end{tabular}
\end{table}

Each game is trained independently (single-task), with 2 evaluator environments running in parallel.

\subsubsection{Additional Training Curves}

\begin{figure}[htbp]
    \centering
    \includegraphics[width=\textwidth]{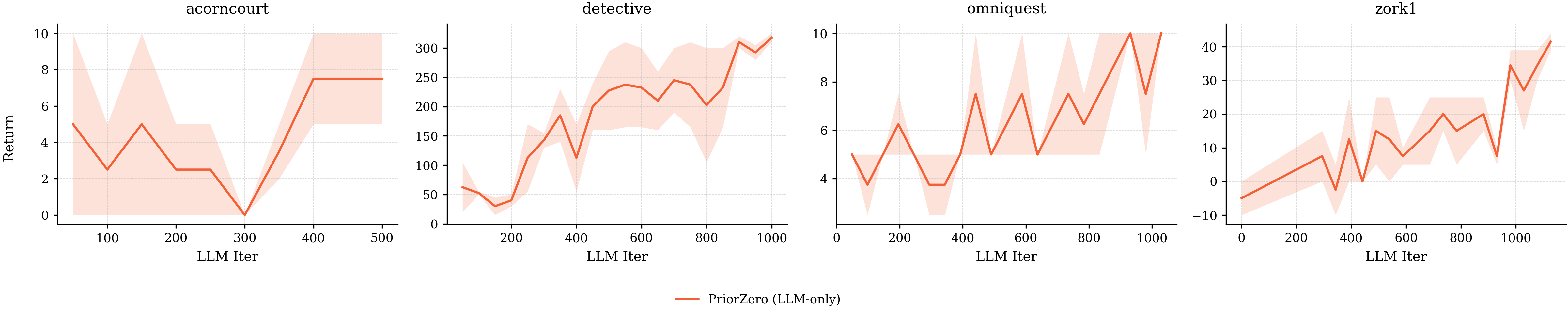}
    \caption{Standalone LLM policy performance during PriorZero's alternating training across four Jericho environments. The LLM policy quality improves progressively with each alternating fine-tuning cycle, demonstrating that the world model's value estimates provide effective signals for LLM adaptation. Periodic dips correspond to the beginning of new fine-tuning cycles before the LLM re-adapts.}
    \label{fig:main_performance_llm}
\end{figure}

During alternating training, the LLM module itself undergoes progressive improvement. Figure~\ref{fig:main_performance_llm} shows the LLM's standalone policy performance across training, illustrating how the world model's value estimates serve as effective supervision signals for LLM adaptation.

\subsubsection{Root-Prior Injection Analysis}

\begin{figure}[htbp]
    \centering
    \includegraphics[width=\textwidth]{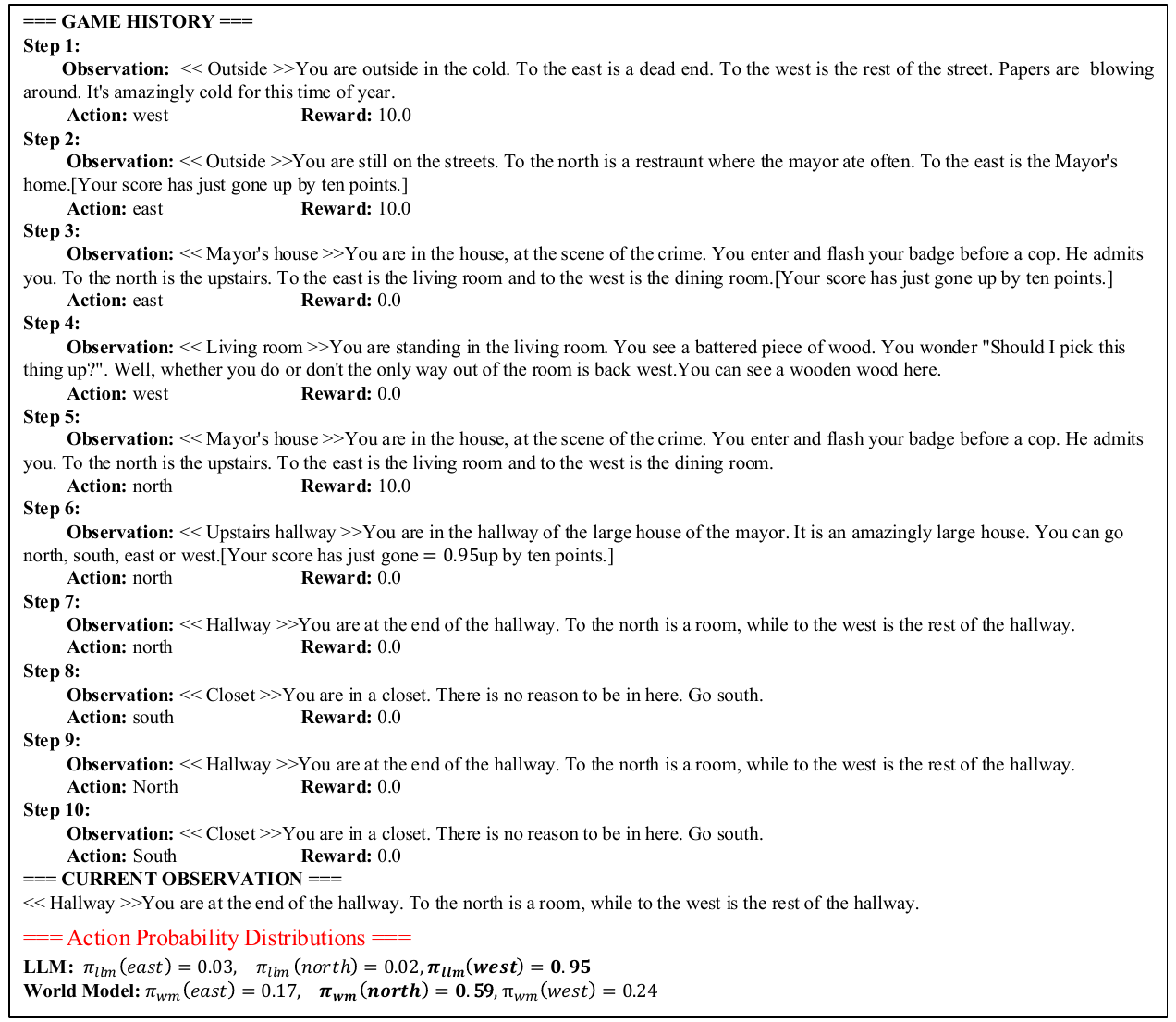}
    \caption{Case study of root-node prior fusion. The LLM assigns high probability to \texttt{west}, while the world model still favors \texttt{north}; fusion redirects MCTS toward the action that breaks the north--south dead loop. }
    \label{fig:frozen_prior_case}
\end{figure}

Figure~\ref{fig:main_entropy}
compares the MCTS root-node visit distribution entropy between PriorZero and UniZero during training, and Figure~\ref{fig:frozen_prior_case} presents qualitative studies of the prior fusion process at the root node.

\paragraph{Two-phase entropy dynamics.}
The entropy curves in Figure~\ref{fig:main_entropy} exhibit a characteristic two-phase pattern that goes beyond a simple ``lower entropy $=$ better search'' interpretation. In the \emph{early phase}, PriorZero's root-node entropy is lower than UniZero's: the LLM prior provides an informative initial search direction, allowing the agent to concentrate visits on semantically meaningful actions instead of repeatedly revisiting low-value branches. In the large admissible-action space of \texttt{zork1}, such directional guidance is crucial for avoiding inefficient or cyclic exploration before the world model has learned reliable value estimates. As training proceeds, PriorZero maintains comparable or higher returns while its root-node entropy rises \emph{above} UniZero's. This reversal indicates that root-prior fusion does not collapse search onto the most likely action; rather, the LLM prior guides MCTS toward semantically plausible regions while still leaving sufficient room to compare alternative actions through world-model rollouts. In this sense, root-prior injection improves exploration \emph{quality}: it filters out obviously unproductive directions early, and later prevents the search distribution from being dominated by brittle world-model preferences. In the final phase of training (last 20\% of evaluations), PriorZero attains a substantially higher mean return than UniZero on \texttt{zork1}.

\paragraph{Action-level case study.}
Figure~\ref{fig:frozen_prior_case} illustrates this effect at the action level. In the game history, the agent repeatedly enters a closet to the north and returns south, forming an unproductive loop, while the current observation indicates that moving west leads back to the hallway. The LLM assigns most probability mass to \texttt{west}, whereas the world model still favors \texttt{north}. Root-prior fusion therefore redirects MCTS toward the appropriate action and helps break the dead-loop exploration pattern. More generally, these examples show that LLM priors can provide useful initial directions for MCTS, especially when the learned world model has not yet distinguished productive actions from locally repetitive but low-value behavior.

\subsection{BabyAI Details}
\label{sec:appendix_babyai}

BabyAI tests whether PriorZero generalizes beyond text-based games. Compared with Jericho, BabyAI differs in three dimensions that interact with our method: it provides \emph{structured} natural-language observations (rather than free-form text), \emph{high-level semantic actions} (rather than verb-object commands), and a \emph{continuous shaped reward} (rather than sparse integer scores). It is also a multi-task benchmark---a single agent is trained across 18 levels simultaneously, in contrast to Jericho where each game is trained independently. The remainder of this subsection details the environment interface, the level set, BabyAI-specific implementation choices, and per-level results.

\subsubsection{Environment and Platform}

BabyAI~\cite{chevalier2019babyai} is a grid-world platform built on MiniGrid that serves as a representative setting for embodied tasks. It features procedurally generated rooms with colored objects (balls, keys, boxes, doors) and natural-language mission specifications. Agents navigate rooms, move objects, unlock doors, and interact with the environment in response to textual commands. We use the AgentGym~\cite{xi2025agentgym} HTTP server to interface with BabyAI environments. The server exposes a REST API (\texttt{/create}, \texttt{/reset}, \texttt{/step}, \texttt{/close}) that manages multiple concurrent environment instances. Each \texttt{/reset} call accepts a \texttt{data\_idx} parameter encoding both the level identity and random seed: $\text{level\_id} = \text{data\_idx} \bmod 40 + 1$, $\text{seed} = \lfloor \text{data\_idx} / 40 \rfloor$.

\paragraph{Observation Space.}
The AgentGym server converts BabyAI's raw $7{\times}7{\times}3$ grid observation into a structured natural-language description. The description includes: (1) visible objects with their colors, types, and relative positions (e.g., ``There is a red ball 1 right in front of you 3 steps away''); (2) wall/barrier distances; (3) the agent's carrying status; and (4) the mission goal (e.g., ``Your goal: go to the red ball''). This text is tokenized by a BGE-base-en-v1.5 encoder (max sequence length 512) for the world model, and presented as-is to the LLM.

\paragraph{Action Space.}
We use the server's \textbf{high-level semantic actions}, which abstract over low-level movement primitives (turn left/right, move forward, pickup, drop, toggle). The server computes reachable actions via BFS path planning at each step, producing a dynamic action space of 3--15 actions such as ``go to red ball 1'', ``pickup blue key 1'', ``toggle and go through green closed door 1''. This is analogous to Jericho's \texttt{valid\_actions} but with smaller branching factor. The maximum action space size is padded to 20 for the world model's fixed-size policy head.

\paragraph{Reward Structure.}
BabyAI uses a continuous reward $r = 1 - 0.9 \cdot (t / T_{\max})$ upon task completion, where $t$ is the number of steps taken and $T_{\max}$ is the maximum episode length. Failed episodes receive $r = 0$. This provides denser feedback than Jericho's sparse integer scores, but still requires efficient planning since reward decays with step count.

\subsubsection{Task Selection}

Following AgentGym-RL~\cite{xi2025agentgymrl}, we train and evaluate on 18 out of 40 BabyAI levels. Table~\ref{tab:babyai_levels} lists all 18 levels grouped by task category.

\begin{table}[htbp]
\centering
\caption{The 18 BabyAI levels used for training and evaluation, following the AgentGym-RL protocol. Levels span four task categories of increasing compositional complexity.}
\label{tab:babyai_levels}
\small
\begin{tabular}{clp{7.5cm}}
\toprule
\textbf{ID} & \textbf{Level Name} & \textbf{Description} \\
\midrule
\multicolumn{3}{l}{\textit{Navigation (11 levels)}} \\
1  & GoToRedBallGrey    & Navigate to a red ball in a grey room \\
2  & GoToRedBall        & Navigate to a red ball with distractors \\
3  & GoToRedBallNoDists & Navigate to a red ball without distractors \\
4  & GoToObjS6          & Navigate to a named object in a 6$\times$6 room \\
5  & GoToLocalS8N7      & Navigate to a local object (8$\times$8, 7 objects) \\
6  & GoToObjMazeS7      & Navigate to an object through a 7$\times$7 maze \\
7  & GoToImpUnlock      & Navigate requiring implicit door unlocking \\
8  & GoToSeqS5R2        & Sequential navigation to two objects \\
9  & GoToRedBlueBall    & Navigate to a red or blue ball \\
10 & GoToDoor           & Navigate to a specified door \\
11 & GoToObjDoor        & Navigate to an object or door by description \\
\midrule
\multicolumn{3}{l}{\textit{Pickup (3 levels)}} \\
19 & PickupLoc          & Pick up an object specified by location \\
20 & PickupDistDebug    & Pick up an object with distractors (debug) \\
21 & PickupAbove        & Pick up an object above another object \\
\midrule
\multicolumn{3}{l}{\textit{Interaction (2 levels)}} \\
30 & ActionObjDoor      & Perform a specified action on an object or door \\
31 & FindObjS7          & Find a hidden object across multiple rooms \\
\midrule
\multicolumn{3}{l}{\textit{Compositional (2 levels)}} \\
33 & OneRoomS20         & Complete a task in a large 20$\times$20 single room \\
36 & SynthLoc           & Synthesized multi-step task with location constraints \\
\bottomrule
\end{tabular}
\end{table}

\subsubsection{Implementation and Configuration}

Table~\ref{tab:babyai_vs_jericho} lists the key hyperparameters and environment settings used in the Jericho and BabyAI experiments.

\begin{table}[htbp]
\centering
\caption{Hyperparameters and environment settings for the Jericho and BabyAI experiments.}
\label{tab:babyai_vs_jericho}
\small
\begin{tabular}{lcc}
\toprule
\textbf{Parameter} & \textbf{Jericho} & \textbf{BabyAI} \\
\midrule
LLM model           & Qwen2.5-3B-Instruct & Qwen2.5-7B-Instruct \\
Environment interface & Local Python API    & HTTP (AgentGym server) \\
Action space size    & 10--100+            & 3--15 (high-level) \\
Max episode steps    & 50--500             & 20 \\
Reward type          & Sparse integer      & Continuous $[0, 1]$ \\
KL coefficient       & 0.01                & 0.1 \\
Entropy bonus        & 0.0                 & 0.001 \\
WM update iters      & 2000                & 500 \\
LLM update iters     & 200                 & 100 \\
MCTS sims (collect)  & 25                  & 25 \\
MCTS sims (eval)     & 25                  & 50 \\
Fusion weight $\alpha$ & 0.5              & 0.5 \\
CoT loss weight $w_{\mathrm{cot}}$  & 0.1                 & 0.1 \\
LLM learning rate    & $1 \times 10^{-6}$  & $1 \times 10^{-6}$ \\
Train batch size     & 128                 & 128 \\
Micro train batch size & 4                 & 2 \\
Replay buffer size   & $3 \times 10^5$     & $5 \times 10^5$ \\
WM batch size        & 64                  & 64 \\
WM learning rate     & $3 \times 10^{-4}$  & $3 \times 10^{-4}$ \\
Text encoder         & BGE-base-en-v1.5    & BGE-base-en-v1.5 \\
\# training tasks    & 4 (single-task)     & 18 (multi-task) \\
\bottomrule
\end{tabular}
\end{table}

\paragraph{LLM Inference Configuration.}
Table~\ref{tab:llm_inference} summarizes the LLM inference parameters. Both environments use vLLM~\cite{kwon2023efficient} for batched inference with DeepSpeed ZeRO-2 for training.

\begin{table}[htbp]
\centering
\caption{LLM inference and training infrastructure parameters.}
\label{tab:llm_inference}
\small
\begin{tabular}{lcc}
\toprule
\textbf{Parameter} & \textbf{Jericho} & \textbf{BabyAI} \\
\midrule
\multicolumn{3}{l}{\textit{vLLM Inference}} \\
Sampling temperature     & 1.0   & 1.0 \\
Top-$p$                  & 0.95  & 0.95 \\
Prompt max length        & 8192  & 4096 \\
Generate max length      & 512   & 512 \\
Log-prob reduction       & mean  & mean \\
Tensor parallel size (3B)  & 1   & --- \\
Tensor parallel size (7B)  & 2   & 2 \\
GPU memory utilization (3B) & 0.25 & --- \\
GPU memory utilization (7B) & 0.35 & 0.35 \\
Prefix caching           & off   & off \\
\midrule
\multicolumn{3}{l}{\textit{CoT Generation}} \\
CoT temperature          & 1.0   & 1.0 \\
CoT top-$p$              & 1.0   & 1.0 \\
CoT max tokens           & 512   & 512 \\
CoT stop tokens          & \texttt{"\textbackslash n\textbackslash n"} & \texttt{"\textbackslash n\textbackslash n"} \\
CoT enabled by default   & Yes    & Yes \\
\midrule
\multicolumn{3}{l}{\textit{DeepSpeed Training}} \\
ZeRO stage               & 2     & 2 \\
Gradient clipping        & 1.0   & 1.0 \\
Adam $(\beta_1, \beta_2)$ & (0.9, 0.95) & (0.9, 0.95) \\
Weight decay             & 0.01  & 0.01 \\
LR scheduler             & cosine w/ min & cosine w/ min \\
LR warmup ratio          & 0.03  & 0.03 \\
PPO clip $(\epsilon_{\text{low}}, \epsilon_{\text{high}})$ & (0.2, 0.2) & (0.2, 0.2) \\
KL estimator             & k3    & k3 \\
Format reward weight     & 0.5   & 0.3 \\
Advantage normalization  & global batch & global batch \\
\bottomrule
\end{tabular}
\end{table}

\paragraph{Action Prior Extraction Pipeline.}
The action prior extraction does \emph{not} parse free-form LLM output. Instead, it uses vLLM's \texttt{prompt\_logprobs} feature to compute the conditional log-probability of each candidate action given the prompt context. For each valid action $a_i$ in the current state:

\begin{enumerate}[leftmargin=*,nosep]
    \item \textbf{Prompt construction}: The system prompt and user prompt are concatenated via the tokenizer's chat template. If CoT is enabled, a CoT prefix (generated in a separate vLLM call) is prepended to the action text.
    \item \textbf{Full sequence assembly}: The prompt token IDs are concatenated with the label token IDs: $[\text{prompt\_ids}; \text{label\_ids}]$, where $\text{label\_ids}$ encodes either ``\texttt{Action: <action>}<eos>'' (no CoT) or ``\texttt{<cot\_prefix> <action>}<eos>'' (with CoT).
    \item \textbf{Scoring}: All sequences are submitted to vLLM in a single batch with \texttt{prompt\_logprobs=1} and \texttt{max\_tokens=1}. The per-token log-probabilities of the label portion are extracted and averaged (\texttt{reduction=mean}) to produce the action score: $s(a_i) = \frac{1}{|L_i|} \sum_{j \in L_i} \log p(t_j \mid t_{<j})$.
    \item \textbf{Prior distribution}: The scores $\{s(a_i)\}$ form the LLM prior in log-probability space. Temperature scaling ($T_{\text{Jericho}}{=}1.0$, $T_{\text{BabyAI}}{=}1.0$) is applied before fusion with the world model's policy logits at the MCTS root node.
\end{enumerate}

This prompt-logprobs approach avoids the fragility of regex-based output parsing and provides calibrated probability estimates over the full action vocabulary.

\paragraph{Format Reward.}
When CoT is enabled, a binary format reward $r_{\mathrm{fmt}}\in\{0,1\}$ is set to 1 iff the LLM output strictly matches \texttt{Reasoning: <analysis>}\textbackslash n\texttt{Action: <action>} with exactly one occurrence of each keyword. Following the blending introduced in \S\ref{subsec:decoupled_training}, the final advantage is $\hat{A}_{\mathrm{final}} = (1-\lambda_{\mathrm{fmt}})\,\hat{A} + \lambda_{\mathrm{fmt}}\,r_{\mathrm{fmt}}$, with $\lambda_{\mathrm{fmt}}{=}0.5$ for Jericho and $\lambda_{\mathrm{fmt}}{=}0.3$ for BabyAI.

\paragraph{Multi-Task Training.}
Unlike Jericho where each game is trained independently, BabyAI follows the AgentGym-RL multi-task protocol: a single agent is trained across all 18 levels simultaneously. During collection, levels are sampled uniformly at random; during evaluation, all 18 levels are cycled deterministically to ensure complete coverage.

\paragraph{LLM Prompt Design.}
The BabyAI-specific system prompt (Figure~\ref{fig:prompt_babyai}) instructs the LLM as ``an expert agent navigating a BabyAI grid-world environment'' and enumerates the six available high-level action types: basic movement (turn left/right, move forward), navigation (\texttt{go to <obj> <id>}), object interaction (\texttt{pick up <obj> <id>}), and door operations (\texttt{go through}, \texttt{toggle and go through}, \texttt{toggle}). The user prompt includes the action history with rewards (\texttt{history\_with\_reward=True}), the current observation, and the list of valid actions (\texttt{observation\_with\_valid\_actions=True}). When CoT is enabled (default for BabyAI), the LLM produces a ``Reasoning'' section analyzing the observation and goal before outputting the chosen action.



\paragraph{Baselines.}
The UniZero baseline uses the identical world model architecture (2-layer transformer, $d{=}768$, 24 heads) and MCTS configuration but without any LLM module. All world-model hyperparameters are kept identical to PriorZero for a fair ablation. 

\paragraph{Hardware and Wall-Clock Time.}
All experiments are run on NVIDIA H200 GPUs.
For \textbf{Jericho}, each single-task run uses 4 H200 GPUs (4-way DDP) with the Qwen2.5-3B-Instruct backbone; one full run reaches $50\text{k}$ environment steps in approximately $35$ wall-clock hours.
For \textbf{BabyAI-18}, each multi-task run uses 4 H200 GPUs (2-way DDP combined with vLLM tensor parallelism for the Qwen2.5-7B-Instruct model); one full run reaches $100\text{k}$ environment steps in approximately $48$ wall-clock hours.
The higher per-step cost on BabyAI is dominated by the larger 7B LLM and the multi-task setting (18 levels sampled per iteration); the higher absolute step count required by BabyAI partially offsets the per-step gain.

\paragraph{Computational cost comparison.}                                                                                                          Table~\ref{tab:compute_comparison} summarizes the per-iteration compute profile of each paradigm under equivalent environment budgets.           
  \begin{table}[h]
  \centering
  \footnotesize
  \caption{Approximate per-iteration compute breakdown on Jericho (Qwen2.5-3B, 4$\times$GPU).}
  \label{tab:compute_comparison}
  \begin{tabular}{lccc}
  \toprule
  \textbf{Component} & \textbf{Naive RLFT (P2)} & \textbf{PriorZero} & \textbf{UniZero} \\
  \midrule
  LLM forward (rollout)  & Every step & Root only & --- \\
  LLM backward (training) & Every iteration & Every $K$-th WM iter & --- \\
  WM forward/backward     & --- & Every iteration & Every iteration \\
  Reference model memory   & Yes (frozen copy) & Yes (frozen copy) & --- \\
  \bottomrule
  \end{tabular}
  \end{table}

  PriorZero's LLM compute is concentrated in two places: (1)~a single forward pass
  per MCTS root node during rollout (amortized over the ${\sim}50$ MCTS simulations
  that follow without LLM involvement), and (2)~PPO updates on the LLM every $K$
  world-model iterations during training. Because the alternating schedule performs
  far fewer LLM gradient steps than Naive RLFT (which updates the LLM at every
  iteration), PriorZero's total LLM training cost is a fraction of P2's despite
  using the same model size. The world-model component (2-layer transformer,
  $d{=}768$) adds negligible overhead relative to the LLM.

\subsubsection{Per-Level Evaluation Curves}
\label{subsec:babyai_per_level}

\begin{figure}[htbp]
    \centering
    \includegraphics[width=\textwidth]{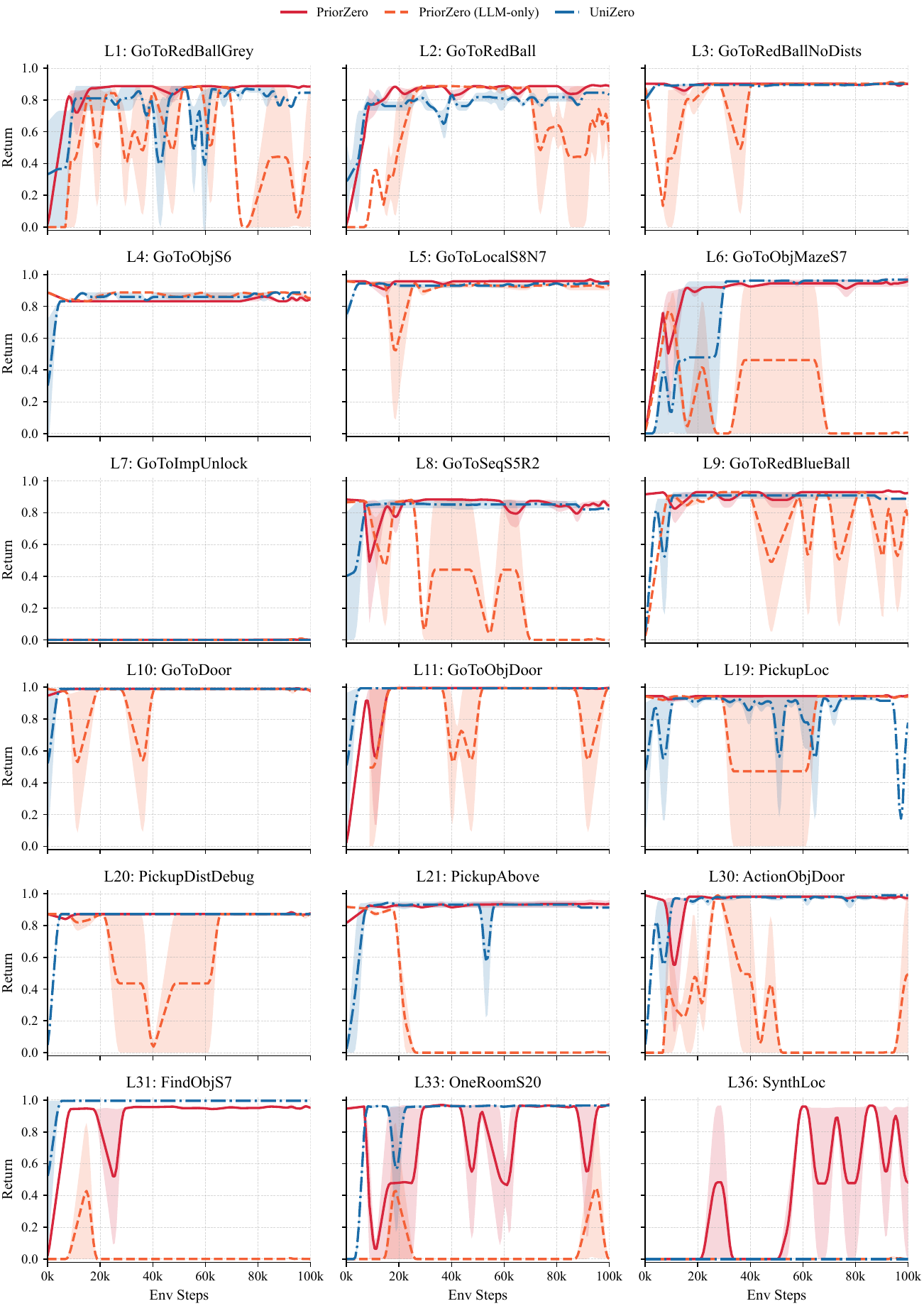}
    \caption{Per-level return on all 18 BabyAI levels. Each subplot shares a common $x$-axis (Env Steps, obtained by the data collector step counter) and $y$-axis (return in $[0,1]$). Curves are means over two seeds; shaded bands span the min--max envelope across seeds. Red: \textbf{PriorZero}. Orange (dashed): \textbf{PriorZero (LLM-only)} — the fine-tuned LLM policy alone. Blue: \textbf{UniZero}. The three categories at the top are Navigation (L1--L11), Pickup (L19--L21), and Interaction/Compositional (L30--L36).}
    \label{fig:babyai_eval_curves}
\end{figure}

\paragraph{Per-Level Curves.}
Fig.~\ref{fig:babyai_eval_curves} presents detailed curves for all 18 BabyAI levels, complementing the aggregated return comparison in the main text (Figure~\ref{fig:babyai_train_reward}). We show a three-way comparison: \textbf{PriorZero} uses the full framework (world model + root-level LLM prior + MCTS); the \textbf{standalone LLM policy} (labeled ``PriorZero (LLM-only)'' in the figure legend) keeps the same fine-tuned LLM module but evaluates with the LLM policy alone (no world-model lookahead / MCTS); and \textbf{UniZero} is the world-model baseline without any LLM prior. This decomposition isolates (i)~how much the world model and planning contribute beyond the LLM's standalone behavior, and (ii)~how much the LLM prior contributes beyond UniZero's planning.

\paragraph{Where Does the World Model Contribute Beyond the LLM?}
Comparing PriorZero against the standalone LLM policy isolates the gain from world-model planning, which scales with task compositionality:
\begin{itemize}[leftmargin=*,nosep]
    \item \textbf{Compositional / multi-room tasks} (\texttt{GoToObjMazeS7}, \texttt{OneRoomS20}, \texttt{ActionObjDoor}, \texttt{FindObjS7}, \texttt{PickupAbove}): the standalone LLM fails outright (return ${\approx}0$), while the full stack reaches ${\ge}0.90$. These tasks require ordered sub-goal sequences that open-loop LLM sampling cannot plan; MCTS lookahead through the learned world model is the decisive signal.
    \item \textbf{Navigation with distractors} (\texttt{GoToRedBallGrey}, \texttt{GoToRedBall}, \texttt{GoToRedBlueBall}): the standalone LLM achieves $0.4$--$0.7$ but is unstable, whereas the full stack sits consistently above $0.88$. The LLM identifies the right object class; world-model value bootstrapping disambiguates among visually similar candidates.
    \item \textbf{Easy single-object tasks} (\texttt{GoToRedBallNoDists}, \texttt{GoToDoor}, \texttt{GoToLocalS8N7}, \texttt{PickupLoc}, \texttt{PickupDistDebug}): the standalone LLM already saturates (${\ge}0.87$); the world model adds at most a few points. The full stack pays no tax on easy levels.
\end{itemize}

\paragraph{Where Does the LLM Prior Help on Top of UniZero?}
Comparing PriorZero against UniZero isolates the gain from injecting the LLM prior at the MCTS root. The most striking case is \texttt{SynthLoc} (L36), where UniZero's mean return collapses to $0.0$ while PriorZero reaches $0.96$: the level uses locative compositional instructions (``next to'', ``to the left of'') that benefit disproportionately from LLM grounding. Smaller gains ($+5$ to $+13$ points) appear on \texttt{GoToRedBallGrey}, \texttt{GoToRedBall}, \texttt{GoToRedBlueBall}, and \texttt{PickupLoc}, where the action space contains near-synonymous candidates that the world model alone cannot cheaply discriminate.

\paragraph{Anomalies.}
Three levels deserve flagging. \texttt{GoToImpUnlock} (L7) requires implicit door unlocking; both UniZero and all PriorZero variants stall near $0$ within $\sim\!100\text{k}$ env steps, suggesting longer training or an explicit intermediate-reward curriculum is needed. \texttt{SynthLoc} (L36) is the clearest case where the LLM prior is \emph{necessary}, not merely accelerating: random exploration never grounds locative phrases, but the frozen LLM vocabulary supplies this grounding for free. \texttt{FindObjS7} (L31) is the only level where UniZero slightly edges out PriorZero ($0.995$ vs.\ $0.960$); both saturate and the gap is within UniZero's seed variation.


\section{Detailed Related Work}
\label{sec:appendix_related_work}
\subsection{Pre-training in Reinforcement Learning}

To address the issues of low sample efficiency and poor generalization in high-dimensional observation spaces, recent studies have explored integrating self-supervised pre-training and representation learning into reinforcement learning (RL). These approaches aim to improve policy learning and generalization through unsupervised or cross-task representations. Early work on World Models~\citep{ha2018world} demonstrated the potential of representation pre-training for improving sample efficiency by learning a generative model of environment dynamics in a latent space and training control policies on top of it. In the visual domain, CURL~\citep{laskin2020curl} leverages contrastive learning to extract visual features from pixel observations, alleviating the representation bottleneck in RL. Seo et al.~\citep{seo2022reinforcement} propose action-free video pre-training, where environment dynamics are first learned from large-scale unlabeled videos and then fine-tuned with action-conditioned world models in downstream tasks, while intrinsic rewards are constructed from latent representations to guide exploration. PIE-G~\citep{yuan2022pre} transfers ImageNet-pretrained encoders to RL and achieves cross-environment zero-shot generalization. VIP~\citep{ma2022vip} formulates representation learning as conditional value function estimation and extracts temporally smooth implicit values from third-person videos to provide dense unsupervised rewards. ContextWM~\citep{wu2023pre} further improves robustness and transferability in real-world video data by explicitly disentangling background and dynamic information through a context encoder. In text-based environments, Dynalang~\citep{lin2024dynalang} trains world models on large-scale text corpora and performs planning and policy optimization in the latent space, enabling task priors without explicit environment interaction. Starling~\citep{basavatia2024starling} leverages large language models (LLMs) to generate self-supervised signals that guide sample-efficient policy learning in text-based environments.

Different from these approaches, \textbf{PriorZero} does not rely on additional representation pre-training or cross-modal datasets. Instead, it directly incorporates LLM priors into the world-model planning process---the world model is trained online from scratch while the LLM prior is adapted in-the-loop, avoiding the granularity mismatch (P3 in \S\ref{sec:motivation}) that arises when LLMs are used as dynamics models.

\subsection{Leveraging LLM Priors in Reinforcement Learning}

With the emergence of large language models (LLMs) that possess strong world knowledge and reasoning capabilities, recent research has explored leveraging them to assist reinforcement learning agents. Li et al.~\citep{li2022pre} first introduced pretrained language models to provide commonsense priors for state understanding and action selection in text-based games. Huang et al.~\citep{huang2022language} propose using LLMs as zero-shot planners that generate executable high-level action sequences from textual prompts to guide embodied agents. SayCan~\citep{ahn2022can} further combines LLM reasoning with value function estimation to improve the success rate of robotic task execution. Beyond planning, LLMs have also been used to guide exploration. The ELLM framework~\citep{cao2024beyond} generates semantically meaningful sub-goals with language models and uses them as exploration rewards to make policy learning more goal-directed. Inner Monologue~\citep{huang2022language} demonstrates that LLM-generated ``internal monologues'' can coordinate multiple modules in embodied systems. DLLM~\citep{liu2024dellma} embeds LLM-generated sub-goals into model-based RL simulations and assigns higher rewards to trajectories consistent with language prompts, thereby improving exploration efficiency in long-horizon tasks. Shi et al.~\citep{shi2025monte} further incorporate LLMs directly into Monte Carlo Tree Search (MCTS), where candidate actions generated by LLMs guide planning in text-based games.

In contrast to these methods---which span paradigms P1 (Naive Policy: \citep{huang2022language,ahn2022can}), P4 (Text Encoder: \citep{li2022pre}), and LLM-guided exploration/search (\citep{cao2024beyond,liu2024dellma,shi2025monte})---\textbf{PriorZero} integrates LLM priors with a learnable world model through root-only injection and alternating fine-tuning. This enables the agent to continuously refine its decision-making policy through environment interaction while keeping dynamics learning uncontaminated by token-level bias.

\subsection{Reinforcement Learning for Fine-Tuning LLMs}

Reinforcement Learning from Human Feedback (RLHF) has become the dominant paradigm for aligning large language models. Christiano et al.~\citep{christiano2017deep} first proposed learning reward models from human preferences and optimizing policies accordingly. Ziegler et al.~\citep{ziegler2019fine} applied this framework to language model summarization, and Stiennon et al.~\citep{stiennon2020learning} further validated its effectiveness on long-form summarization tasks. InstructGPT~\citep{ouyang2022traininglanguagemodelsfollow} extended RLHF to large-scale models with tens of billions of parameters, combining PPO optimization with human preference data to significantly improve instruction-following behavior. To reduce the cost of human annotations, Constitutional AI~\citep{bai2022constitutional} and RLAIF~\citep{lee2023rlaif} replace human feedback with AI-generated preferences or LLM-based reward scoring, achieving competitive performance across multiple tasks. In decision-making scenarios, Reinforcement Learning Fine-Tuning (RLFT) has been explored to improve multi-step decision capabilities of LLMs. Schmied et al.~\citep{schmied2025llmsgreedyagentseffects} systematically study the effects of RLFT on decision-making behaviors and show that RLFT can mitigate greedy tendencies and the ``knowing--doing gap'' in LLMs. However, their experiments are mainly limited to short-horizon tasks such as bandits and tic-tac-toe. Recent works further combine self-play or search mechanisms to generate high-quality training signals. For example, rStar-Math~\citep{guan2025rstar} performs deep exploration in reasoning space via Monte Carlo Tree Search and filters correct reasoning paths using code execution feedback to construct high-quality process supervision data for RFT. STAIR~\citep{zhang2025stair} fine-tunes models on reasoning chains containing introspective reasoning under safety constraints for alignment.

Unlike these methods, \textbf{PriorZero} constructs reinforcement learning advantage signals using value estimates generated during world-model planning. This provides low-variance training feedback that resolves the credit-assignment failure limiting end-to-end RLFT (P2 in \S\ref{sec:motivation}) and allows RLFT to scale from short-horizon decision tasks to complex long-horizon environments.

\section{Prompt Template}
\label{sec:appendix_prompt}

\begin{figure}[htbp]
    \centering
    \includegraphics[width=\textwidth]{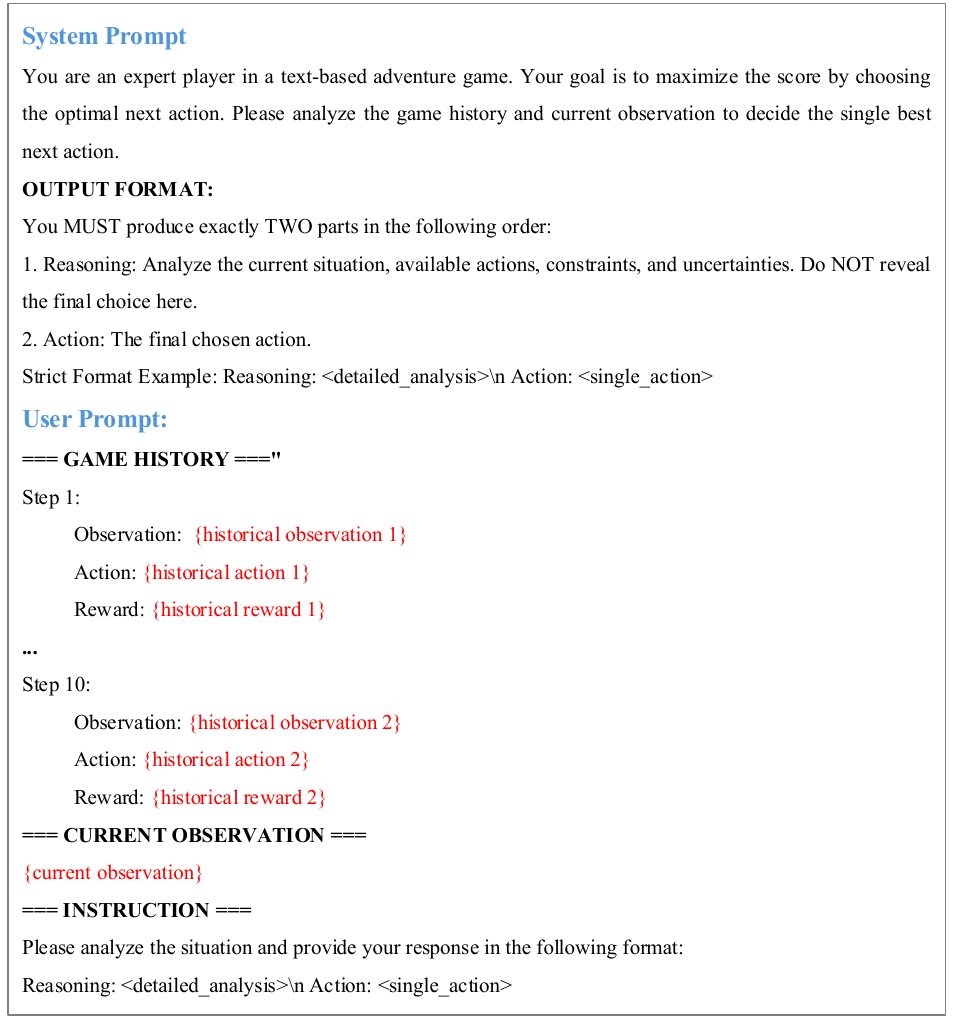}
    \caption{Prompt template for extracting semantic action priors from the LLM in \textbf{Jericho} environments. Red text indicates placeholders filled with task-specific content (game description, current observation, valid actions). The template enforces a two-stage generation: first a neutral situation analysis (CoT reasoning), then a probability distribution over candidate actions.}
    \label{fig:prompt}
\end{figure}

PriorZero uses a unified two-stage prompt template to extract semantic action priors from the LLM during rollout, following the structured prior extraction process described in \S\ref{subsec:root_injection}. The template consists of a \textbf{system prompt} (defining the agent role, environment context, and output format) and a \textbf{user prompt} (injecting the interaction history, current observation, and valid actions). The LLM first produces a neutral situation analysis (CoT reasoning), then outputs a probability distribution over candidate actions.
Both Jericho and BabyAI share the same prompt skeleton and scoring pipeline---the BabyAI variant inherits all LLM-calling, CoT-generation, and prompt-logprobs scoring logic from the Jericho implementation, overriding only the system prompt and user prompt text to reflect the grid-world domain. Neither environment uses few-shot examples; the prompt relies entirely on zero-shot instruction following.
Figure~\ref{fig:prompt} shows the Jericho prompt template, and Figure~\ref{fig:prompt_babyai} shows the BabyAI variant. Red placeholders are filled at each step with the current observation, action history, and valid action list.

\begin{figure}[htbp]
    \centering
    \fbox{\parbox{0.95\textwidth}{\small
    \textbf{System Prompt:}\\[4pt]
    You are an expert agent navigating a BabyAI grid-world environment.
    You are placed in rooms and must accomplish goals by choosing optimal actions.\\[4pt]
    Available action types:\\
    \hspace*{1em}-- turn left / turn right / move forward: basic movement\\
    \hspace*{1em}-- go to \textcolor{red}{\texttt{<obj> <id>}}: navigate to a specific object\\
    \hspace*{1em}-- pick up \textcolor{red}{\texttt{<obj> <id>}}: pick up an object\\
    \hspace*{1em}-- go through \textcolor{red}{\texttt{<door> <id>}}: go through an open door\\
    \hspace*{1em}-- toggle and go through \textcolor{red}{\texttt{<door> <id>}}: open and go through a closed/locked door\\
    \hspace*{2em}(locked doors require a matching color key)\\
    \hspace*{1em}-- toggle: open/close a door directly in front of you\\[4pt]
    Your goal is to complete the given task efficiently to maximize your score.\\[4pt]
    OUTPUT FORMAT:\\
    You MUST produce exactly TWO parts in the following order:\\
    1. Reasoning: Analyze the current observation, your position, nearby objects,
    and which action best progresses toward the goal.\\
    2. Action: The final chosen action (must be one of the valid actions).\\
    Strict Format Example:\\
    Reasoning: \textcolor{red}{\texttt{<detailed\_analysis>}}\\
    Action: \textcolor{red}{\texttt{<single\_action>}}\\[8pt]
    \textbf{User Prompt:}\\[4pt]
    === ACTION HISTORY ===\\
    Step 1:\\
    \hspace*{1em}Observation: \textcolor{red}{\texttt{<obs\_text\_1>}}\\
    \hspace*{1em}Action: \textcolor{red}{\texttt{<action\_1>}}\\
    \hspace*{1em}Reward: \textcolor{red}{\texttt{<reward\_1>}}\\
    \ldots\\[4pt]
    === CURRENT OBSERVATION ===\\
    \textcolor{red}{\texttt{<current\_obs\_text>}}\\[4pt]
    {[Valid Actions]}\\
    Choose from: \textcolor{red}{\texttt{`go to red ball 1', `pick up blue key 1', \ldots}}\\[4pt]
    === INSTRUCTION ===\\
    Analyze the observation and provide your response:\\
    Reasoning: \textcolor{red}{\texttt{<detailed\_analysis>}}\\
    Action: \textcolor{red}{\texttt{<single\_action>}}
    }}
    \caption{Prompt template for extracting semantic action priors from the LLM in \textbf{BabyAI} environments. The template shares the same two-stage structure as the Jericho variant (Figure~\ref{fig:prompt}), with three key differences: (1)~the system prompt enumerates BabyAI's high-level semantic action types; (2)~the user prompt includes valid actions by default (controlled by \texttt{observation\_with\_valid\_actions=True}); (3)~CoT reasoning is enabled by default. Red text indicates placeholders filled at each step.}
    \label{fig:prompt_babyai}
\end{figure}


\newpage
\section*{NeurIPS Paper Checklist}

\begin{enumerate}

\item {\bf Claims}
    \item[] Question: Do the main claims made in the abstract and introduction accurately reflect the paper's contributions and scope?
    \item[] Answer: \answerYes{}
    \item[] Justification: The abstract and introduction accurately describe our contributions: root-prior injection for MCTS guidance and alternating fine-tuning for LLM adaptation, validated on Jericho and BabyAI benchmarks.

\item {\bf Limitations}
    \item[] Question: Does the paper discuss the limitations of the work performed by the authors?
    \item[] Answer: \answerYes{}
    \item[] Justification: We acknowledge that our evaluation is limited to text-based (Jericho) and grid-world (BabyAI) environments with discrete action spaces. The conclusion notes this scope and the commented future-work direction (multimodal embodied settings) reflects an acknowledged limitation.

\item {\bf Theory assumptions and proofs}
    \item[] Question: For each theoretical result, does the paper provide the full set of assumptions and a complete (and correct) proof?
    \item[] Answer: \answerNA{}
    \item[] Justification: This paper does not include theoretical results. Our contributions are algorithmic and empirical.

    \item {\bf Experimental result reproducibility}
    \item[] Question: Does the paper fully disclose all the information needed to reproduce the main experimental results of the paper to the extent that it affects the main claims and/or conclusions of the paper (regardless of whether the code and data are provided or not)?
    \item[] Answer: \answerYes{}
    \item[] Justification: We provide full implementation details in Appendix~\ref{sec:appendix_implementation}, including hyperparameters (Table~\ref{tab:babyai_vs_jericho}), prompt templates (Figures~\ref{fig:prompt} and~\ref{fig:prompt_babyai}), and training procedures. Code will be released upon acceptance.

\item {\bf Open access to data and code}
    \item[] Question: Does the paper provide open access to the data and code, with sufficient instructions to faithfully reproduce the main experimental results, as described in supplemental material?
    \item[] Answer: \answerYes{}
    \item[] Justification: We will release our code upon acceptance. The environments (Jericho, BabyAI) are publicly available. Implementation details are provided in Appendix~\ref{sec:appendix_implementation}.

\item {\bf Experimental setting/details}
    \item[] Question: Does the paper specify all the training and test details (e.g., data splits, hyperparameters, how they were chosen, type of optimizer) necessary to understand the results?
    \item[] Answer: \answerYes{}
    \item[] Justification: All training and evaluation details are provided in Section~\ref{sec:experiments} and Appendix~\ref{sec:appendix_implementation}, including hyperparameters, environment configurations, and LLM inference settings.

\item {\bf Experiment statistical significance}
    \item[] Question: Does the paper report error bars suitably and correctly defined or other appropriate information about the statistical significance of the experiments?
    \item[] Answer: \answerYes{}
    \item[] Justification: Main comparisons (PriorZero vs.\ UniZero, Figure~\ref{fig:main_performance}) report shaded regions showing variation across 3 random seeds. Ablation variants that clearly failed (frozen prior, non-alternating, $w_{\mathrm{cot}}{=}1.0$, CoT-off, MCTS-off) are reported from a single seed, as documented in the footnote in Section~\ref{sec:experiments}.
    \item[] Guidelines:
    \begin{itemize}
        \item The answer \answerNA{} means that the paper does not include experiments.
        \item The authors should answer \answerYes{} if the results are accompanied by error bars, confidence intervals, or statistical significance tests, at least for the experiments that support the main claims of the paper.
        \item The factors of variability that the error bars are capturing should be clearly stated (for example, train/test split, initialization, random drawing of some parameter, or overall run with given experimental conditions).
        \item The method for calculating the error bars should be explained (closed form formula, call to a library function, bootstrap, etc.)
        \item The assumptions made should be given (e.g., Normally distributed errors).
        \item It should be clear whether the error bar is the standard deviation or the standard error of the mean.
        \item It is OK to report 1-sigma error bars, but one should state it. The authors should preferably report a 2-sigma error bar than state that they have a 96\% CI, if the hypothesis of Normality of errors is not verified.
        \item For asymmetric distributions, the authors should be careful not to show in tables or plots symmetric error bars that would yield results that are out of range (e.g., negative error rates).
        \item If error bars are reported in tables or plots, the authors should explain in the text how they were calculated and reference the corresponding figures or tables in the text.
    \end{itemize}

\item {\bf Experiments compute resources}
    \item[] Question: For each experiment, does the paper provide sufficient information on the computer resources (type of compute workers, memory, time of execution) needed to reproduce the experiments?
    \item[] Answer: \answerYes{}
    \item[] Justification: All experiments run on NVIDIA H200 GPUs. Each Jericho run uses 4 H200 GPUs and reaches $50\text{k}$ environment steps in $\sim\!35$ wall-clock hours; each BabyAI-18 run uses 4 H200 GPUs and reaches $100\text{k}$ environment steps in $\sim\!48$ wall-clock hours. Full hardware and timing details are in Appendix~\ref{sec:appendix_implementation}.

\item {\bf Code of ethics}
    \item[] Question: Does the research conducted in the paper conform, in every respect, with the NeurIPS Code of Ethics \url{https://neurips.cc/public/EthicsGuidelines}?
    \item[] Answer: \answerYes{}
    \item[] Justification: Our research conforms to the NeurIPS Code of Ethics. We use publicly available benchmarks and do not involve human subjects or sensitive data.

\item {\bf Broader impacts}
    \item[] Question: Does the paper discuss both potential positive societal impacts and negative societal impacts of the work performed?
    \item[] Answer: \answerNA{}
    \item[] Justification: This is foundational research on RL algorithms. We do not foresee direct negative societal impacts from our method.

\item {\bf Safeguards}
    \item[] Question: Does the paper describe safeguards that have been put in place for responsible release of data or models that have a high risk for misuse (e.g., pre-trained language models, image generators, or scraped datasets)?
    \item[] Answer: \answerNA{}
    \item[] Justification: We do not release models or datasets. We use publicly available LLMs (Qwen2.5) and benchmarks (Jericho, BabyAI).

\item {\bf Licenses for existing assets}
    \item[] Question: Are the creators or original owners of assets (e.g., code, data, models), used in the paper, properly credited and are the license and terms of use explicitly mentioned and properly respected?
    \item[] Answer: \answerYes{}
    \item[] Justification: We cite all assets used: Jericho (MIT License), BabyAI (MIT License), Qwen2.5 (Apache 2.0), BGE-base-en-v1.5 (MIT License), and UniZero/LightZero (Apache 2.0). All are publicly available with permissive open-source licenses.

\item {\bf New assets}
    \item[] Question: Are new assets introduced in the paper well documented and is the documentation provided alongside the assets?
    \item[] Answer: \answerYes{}
    \item[] Justification: We will release code with documentation and an open-source license (Apache 2.0) upon acceptance.

\item {\bf Crowdsourcing and research with human subjects}
    \item[] Question: For crowdsourcing experiments and research with human subjects, does the paper include the full text of instructions given to participants and screenshots, if applicable, as well as details about compensation (if any)?
    \item[] Answer: \answerNA{}
    \item[] Justification: This paper does not involve crowdsourcing or human subjects research.

\item {\bf Institutional review board (IRB) approvals or equivalent for research with human subjects}
    \item[] Question: Does the paper describe potential risks incurred by study participants, whether such risks were disclosed to the subjects, and whether Institutional Review Board (IRB) approvals (or an equivalent approval/review based on the requirements of your country or institution) were obtained?
    \item[] Answer: \answerNA{}
    \item[] Justification: This paper does not involve human subjects research.

\item {\bf Declaration of LLM usage}
    \item[] Question: Does the paper describe the usage of LLMs if it is an important, original, or non-standard component of the core methods in this research? Note that if the LLM is used only for writing, editing, or formatting purposes and does \emph{not} impact the core methodology, scientific rigor, or originality of the research, declaration is not required.
    \item[] Answer: \answerYes{}
    \item[] Justification: LLMs (Qwen2.5-3B/7B) are a core component of our method, providing semantic action priors for MCTS planning. We describe their usage in Section~\ref{sec:priorzero} and Appendix~\ref{sec:appendix_implementation}.

\end{enumerate}

\end{document}